\documentclass[runningheads]{llncs}

\usepackage{graphicx}
\usepackage{amsmath,amssymb} % define this before the line numbering.
\usepackage[htt]{hyphenat}
\usepackage{booktabs}
\usepackage{subcaption}
\usepackage{enumitem}
\usepackage{tabularx}
\newcolumntype{Y}{>{\centering\arraybackslash}X}
\usepackage[figuresleft]{rotating}
\usepackage{xspace}
\usepackage[pagebackref,breaklinks,colorlinks]{hyperref}
\usepackage[width=122mm,left=12mm,paperwidth=146mm,height=193mm,top=12mm,paperheight=217mm]{geometry}

\newcommand{\modelname}{Transframer\xspace}
\newcommand{\im}{\mathbf{I}}
\newcommand{\dctseq}{\vec{d}}
\newcommand{\ann}{\mathbf{a}}

\newcommand{\view}{\mathbf{v}}
\newcommand{\context}{\mathcal{C}}
\newcommand{\contextset}{\left\{ \left(\im_n, \ann_n\right) \right\}_n}
\newcommand{\imcode}{\im^{\text{code}}}
\newcommand{\imrec}{\hat{\im}}
\newcommand{\target}{\text{target}}
\newcommand{\imtarget}{\im^{\target}}
\newcommand{\anntarget}{\ann^{\target}}
\renewcommand{\vec}[1]{\mathbf{#1}}
\renewcommand{\paragraph}[1]{\vspace{4pt}\noindent {\bf #1}\ }

\begin{document}

\pagestyle{headings}
\mainmatter
%******************
\title{Transframer: Arbitrary Frame Prediction with Generative Models} % Replace with your title
\titlerunning{Transframer: Arbitrary Frame Prediction with Generative Models}
\author{Charlie Nash, João Carreira, Jacob Walker, Iain Barr, \\Andrew Jaegle, Mateusz Malinowski*, Peter Battaglia* \\
{\footnotesize*equal senior authorship} \\
{\tt\small charlienash@deepmind.com}
}
\authorrunning{C. Nash et al.}
\institute{DeepMind, London, UK}
%******************
\maketitle

\begin{abstract}
We present a general-purpose framework for image modelling and vision tasks based on probabilistic frame prediction. Our approach unifies a broad range of tasks, from image segmentation, to novel view synthesis and video interpolation. We pair this framework with an architecture we term \modelname, which uses U-Net and Transformer components to condition on annotated context frames, and outputs sequences of sparse, compressed image features. Transframer is the state-of-the-art on a variety of video generation benchmarks, is competitive with the strongest models on few-shot view synthesis, and can generate coherent 30 second videos from a single image without any explicit geometric information. A single generalist Transframer simultaneously produces promising results on 8 tasks, including semantic segmentation, image classification and optical flow prediction with no task-specific architectural components, demonstrating that multi-task computer vision can be tackled using probabilistic image models. Our approach can in principle be applied to a wide range of applications that require learning the conditional structure of annotated image-formatted data.
\end{abstract}

%%%%%%%%% BODY TEXT
\section{Introduction}
\label{sec:intro}

\begin{figure}[t]
    \centering
    \includegraphics[width=0.9\linewidth]{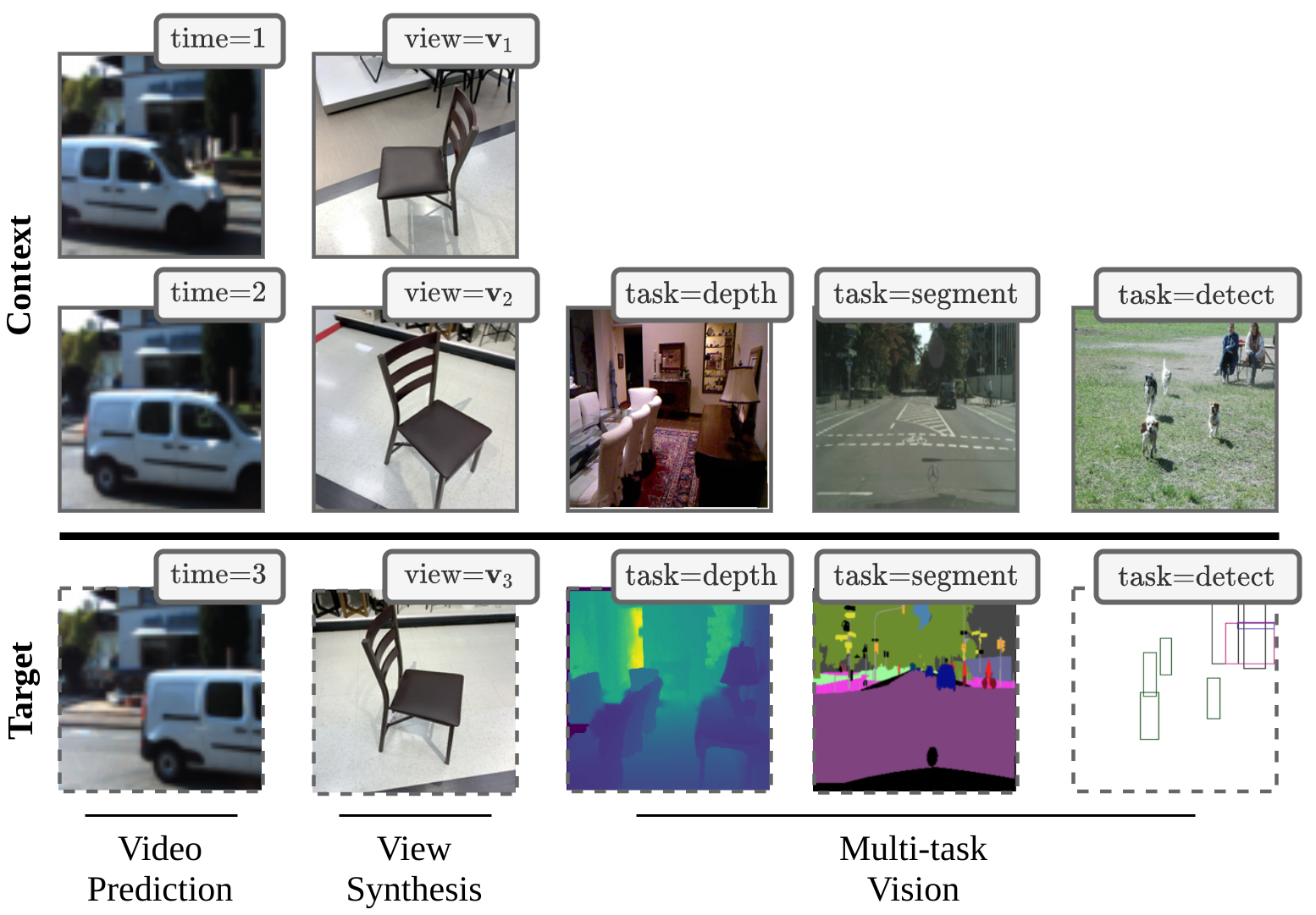}
    \caption{A framework for general visual prediction. Given a collection of context images with associated annotations (time-stamps, camera viewpoints, etc.\ ), and a query annotation, the task is to predict a probability distribution over the target image. This framework supports a range of visual prediction tasks, including video modelling, novel view synthesis, and multi-task vision.}
    \label{fig:teaser}
\end{figure}

We introduce a framework that unifies a wide range of tasks under the umbrella of conditional frame prediction, and demonstrate its efficacy on an extensive range of challenging image modelling and computer vision tasks.

Our proposed model, which we call \emph{\modelname}, is trained to predict the probability of arbitrary frames conditioned on one or more annotated context frames. Context frames could be previous video frames, along with time annotations, or views of a scene with associated camera annotations. \modelname is a likelihood-based autoregressive method analogous to WaveNet~\cite{oord2016wavenet} and large-scale language models~\cite{brown2020language}, implemented with Transformers~\cite{vaswani2017attention} and U-Nets~\cite{ronneberger2015u}, whose underlying representation of the data uses the sparse discrete cosine transform (DCT) image representation introduced by Nash et al.\ \cite{nash2021generating}.

Unlike deterministic methods for frame-to-frame modelling~\cite{eigen2015predicting,ubernet}, \modelname maps context frames to a probability distribution, which allows uncertainty to be quantified and diverse outputs to be obtained through sampling. While autoregressive models are typically expensive or prohibitive to evaluate and sample from, \modelname's sparse representations allow more efficient computation than raw pixels, which we demonstrate by sampling coherent 30 second-long videos.

We applied \modelname to a wide range of tasks, ranging from video modelling, novel view synthesis, semantic segmentation, object recognition, depth estimation, optical flow prediction, etc., and found promising results in all cases. 

Our work makes the following key contributions:
\begin{itemize}[noitemsep,topsep=4pt,label=$\boldsymbol{\cdot}$]
    \item \modelname's quantitative performance is state-of-the-art on video modelling and competitive on novel view synthesis. 
    \item We introduce residual sparse DCT representations as a basis for generative video modelling.
    \item \modelname is general-purpose and can be trained on a wider variety of modelling tasks than are traditionally tackled by more domain-specific methods.
    \item We show how to generate long video sequences (30 seconds) using our probabilistic framework.
\end{itemize}

\section{Background}
\label{sec:background}
To contextualize this work, we first provide a description of the prior work we build on, focusing on generative models of compressed image representations, and the DCTransformer generative image model~\cite{nash2021generating}. An additional discussion of related work on generative video modelling, view synthesis, and alternative frameworks for image-to-image translation can be found in the Appendix A.

\paragraph{Generative modelling with compressed image representations}
\label{subsec:compressed}
To deal with the high dimensionality of image, audio and video data, a number of generative models operate on compressed data representations~\cite{DBLP:conf/nips/OordVK17,DBLP:conf/cvpr/EsserRO21,nash2021generating}. This is particularly important for likelihood-based generative models, which can otherwise waste their capacity on perceptually unimportant details~\cite{DBLP:conf/nips/OordVK17}. For images $\im$ the general approach is as follows:
\begin{enumerate}[noitemsep,topsep=4pt]
    \item Convert images to compressed codes using an encoder \mbox{$\imcode = \text{Encoder}(\im)$}, 
    \item Train a generative model $p(\imcode)$ on the image codes,
    \item Decode samples $\imcode_s \sim p(\imcode)$ from the model back to images $\imrec_s = \text{Decoder}(\imcode_s)$.
\end{enumerate}

\noindent VQ-VAE~\cite{DBLP:conf/nips/OordVK17}, the most well-known of these approaches, uses a vector-quantized autoencoder network to spatially downsample and compress image data, reducing the size of the image representation while maintaining the most important perceptual details. The resulting image codes are modelled post-hoc using a PixelCNN-style autoregressive model, resulting in samples that are more globally coherent than samples from models trained on raw pixel data. VQ-GAN~\cite{DBLP:conf/cvpr/EsserRO21} uses an adversarially trained decoder to achieve higher compression rates and image reconstructions with improved perceptual quality. Transformer-based autoregressive architectures are an effective alternatives to PixelCNN, forming the basis of prior generative models in a number of works~\cite{DBLP:conf/cvpr/EsserRO21,DBLP:journals/corr/abs-2111-12417}. 

\paragraph{Sparse DCT image representations} In this work we use  the DCT-based image representation proposed by Nash et al.\ \cite{nash2021generating}, rather than raw pixels. The DCT-based image representation 
% borrows heavily from 
is heavily inspired by the JPEG codec~\cite{DBLP:journals/cacm/Wallace91}; and it encodes data in several stages:
\begin{enumerate}[noitemsep,topsep=4pt]
    \item The DCT is applied to non-overlapping image patches for each channel in the YUV color space. The UV colour components are optionally 2x downsampled before applying the block-DCT.
    \item The resulting DCT data are lossily quantized using a quality-parameterized quantization matrix. Lower quality settings result in stronger quantization, and a greater loss of information. 
    \item The quantized DCT data are converted to a sparse representation: a list consisting of tuples $\dctseq = (c, p, v)$ of DCT channels, positions, and values for the non-zero elements.
    \item The list is sorted for the purposes of generative modelling. As in Nash et al.\ we order from low-to-high frequency DCT channels, interleaving luminance and color channels. Within DCT channels, list elements are sorted by spatial position in a raster-scan order. 
\end{enumerate}
\noindent The resulting image representation has a number of key properties. It consists of variable-length sequences $\left[\dctseq_l \right]_{l=1}^L$, where the length $L$ depends on the content of the input image. Images with more higher-frequency content result in longer sequences. This contrasts with the compressed representations of VQ-VAE and VQ-GAN, which are fixed size regardless of image content. In addition, the size of the representation and quality of image reconstructions is controllable via the quantization quality setting. We can choose low-quality settings for more efficient training and inference, or high-quality settings for increased image quality. 

\paragraph{DCTransformer} To model the above data representation, Nash et al.\ ~\cite{nash2021generating} proposed the DCTransformer architecture. DCTransformer is an autoregressive model that predicts elements of sparse DCT sequences conditioned on all prior elements:
\begin{align}
    p(\left[\dctseq_l \right]_l) &= \prod_l p(\dctseq_l \ | \ \dctseq_{<l}) \\
    p(\dctseq_l \ | \ \dctseq_{<l}) &= p(v_l | c_l, p_l, \dctseq_{<l})p(p_l | c_l, \dctseq_{<l})p(c_l | \dctseq_{<l})
\end{align}
The architecture has an encoder-decoder structure, where the encoder processes arbitrarily long sequences of DCT data, and the decoder autoregressively predicts fixed-size chunks of DCT data, conditioned on the encoder's outputs. 

During training a target chunk $\left[\dctseq_l\right]_{l=t}^{t+C}$ of length $C$ is randomly selected from the DCT sequence, and all prior elements in the sequence $\left[\dctseq_l\right]_{l=1}^{t}$ are treated as inputs. The sparse inputs are converted to a dense \textit{DCT image}. DCT images are 3D arrays, where each spatial location corresponds to a pixel block of size $B{\times}B$, and channels correspond to different DCT components.  For images of size $H{\times}W{\times}3$ and DCT block size $B$ the corresponding DCT images will therefore be of size $H/B{\times}W/B{\times}3B^2$. Input DCT images are \textit{partially observed} as they consist only of data from the partial DCT input sequence $\left[\dctseq_l\right]_{l=1}^{t}$, rather than the full data sequence. The use of DCT images enables the encoder to process arbitrarily long input sequences, as regardless of the length they are converted to fixed size DCT image arrays. This contrasts with typical sequence-modelling approaches, where memory and computation scales with sequence length.

A Vision-Transformer encoder processes partial DCT images, and a sequence of output embeddings is passed to a causally-masked Transformer decoder. The Transformer decoder embeds and processes target sequences $\left[\dctseq_l\right]_{l=t}^{t+C}$, and uses channel, position and value heads to output predictions for the corresponding list elements. The list elements are treated as discrete values, and predicted using a softmax distribution, with an autoregressive cross-entropy training objective. Writing $\theta_c(\cdot)$, $\theta_p(\cdot)$ and $\theta_v(\cdot)$ for the channel, position and value softmax parameters respectively the objective for target chunk$\left[\dctseq_l\right]_{l=t}^{t+C}$ is:
\begin{align}
    \mathcal{L} = \sum_{l=t}^{t + C}
    \log{\theta_v(c_l, p_l, \dctseq_{<l})} \cdot \vec{v}_l + \log{\theta_p(c_l, \dctseq_{<l})} \cdot \vec{p}_l + 
    \log{\theta_c(\dctseq_{<l})} \cdot  \vec{c}_l
    % \left[\log{\theta_v(c_l, p_l, \dctseq_{<l})}\right]_{v_l} + \left[\log{\theta_p(c_l, \dctseq_{<l})}\right]_{p_l} + \left[\log{\theta_c(\dctseq_{<l})}\right]_{c_l},
    % \log{\theta_c(\dctseq_{<l})} \dot \vec{c}_l,
\end{align}
where $\vec{v}_l$, $\vec{p}_l$ and $\vec{c}_l$ are one-hot vectors associated with indices $v_l$, $p_l$ and $c_l$ respectively.
% where $[x]_a$ is an indicator function equals to one if and only if $x$ is $a$.
In this work we use the same decoder and training objective, but augment the encoder to additionally process annotated context images. 

\section{Methods}
\label{sec:methods}
Here, we present two main components of our work. (1) A framework for probabilistic vision that generalizes video prediction, few-shot view synthesis, and other computer vision tasks, (2) An architecture and modelling paradigm that implements that framework called \modelname, which builds on the DCTransformer~\cite{nash2021generating} generative image model and data representation.

\subsection{Generalized conditional image modelling}

\begin{table*}[t]
    \centering
    \caption{Target, context, and annotation specification for different task types.}
    \resizebox{\linewidth}{!}{
        \begin{tabular}{  m{2cm}  m{1.3cm}  m{2.4cm}  m{8.5cm} }
            \toprule
            \textbf{Task type} & $\anntarget$ & $\context$ & \textbf{Description} \\ \midrule
            Video \mbox{modelling} & $t_{\target}$ & $\left\{ (\im_t, t) \right\}_{t=1}^{t_{\target}-1}$ & Query an image at
            desired time, $t_{\target}$. The $\context$ contains the sequence of $t_{\target}-1$ previous images and associated timestamps, $t$. \\ \midrule
            Video \mbox{interpolation} & $t_{\target}$ & $\left\{ (\im_t, t) \right\}_{t=t_0}^{t_{\target}-1} $\newline$ \cup \left\{ (\im_T, T) \right\}$ & Query an image at desired time, $t_{\target}$. The $\context$ contains a sequence of immediately preceding frames starting at time, $t_0 < t_{\target}$, a future frame at time, $T > t_{\target}$, and associated timestamps, $t$. \\ \midrule
            View \mbox{synthesis} & $\view_{\target}$ & $\left\{ (\im_n, \view_n) \right\}_{n=1}^N$ & Query an image from a desired viewpoint, $\view_{\target}$. The $\context$ contains $N$ other images and associated viewpoints, $\view_n$. \\ \midrule
            \mbox{Multi-format} image \mbox{translation} & $f_{\target}$ & $\left\{ (\im_{f_{\text{RGB}}}, f_{\text{RGB}}) \right\}$ & Query a desired image format, $f_{\target}$, e.g., RGB, depth channel, semantic segmentation, etc. The $\context$ contains an RGB image, annotated $f_{\text{RGB}}$. \\ \bottomrule
        \end{tabular}
    }
    \label{tab:taskmodels}
\end{table*}

Our probabilistic formulation is as follows: given an annotation $\anntarget$, and context information $\context$, the task is to predict a distribution over target image $\imtarget$,
\begin{align}
 p\left(\imtarget \ | \ \anntarget, \ \context \right), \label{eq:preddist}
\end{align}
where $\context{=}\contextset$ is a set of %$N$ 
context images $\im_n$ and associated annotations $\ann_n$. An annotation is metadata about the image such as video timestamps, camera coordinates, image channels, etc. This framework can be applied to a wide range of frame-prediction tasks, as long as we can phrase them in terms of $\anntarget$ and $\context$. Some of the tasks we consider, such as novel view synthesis and video generation have previously been expressed in this manner~\cite{eslami2018neural,DBLP:journals/corr/abs-1807-02033}. Other tasks, including semantic segmentation, depth estimation, or optical flow estimation are frequently modeled within the image prediction framework but typically using task-specific losses. Finally, tasks such as object detection and image classification are generally approached using specialized representations and losses, without any image generation components. Our framework enables us to perform all these tasks using a single unified training objective.

In some cases, we want a predictive distribution over multiple target images $\left\{ \imtarget_k \right\}_k$ conditioned on queries $\left\{ \anntarget_k \right\}_k$. For example, to predict multiple consecutive frames of video, or to synthesize views at multiple query viewpoints. The simplest approach is to treat the target images as conditionally independent:
\begin{align}
 p\left(\left\{ \imtarget_k \right\}_k \ | \ \left\{ \anntarget_k \right\}_k, \ \context \right) = \prod_k  p\left(\imtarget_k \ | \ \anntarget_k, \ \context \right),
\end{align}
This enables parallel generation of the target images, but does not capture any additional dependencies between the target images. In cases where the target images are well-constrained by the the context, such as view synthesis with a large collection of input views, this is not a significant problem. However, whenever there exist strong dependencies between target images, as in video modelling for example, sampled output images are likely to be inconsistent with one another. To address this issue, we can instead predict target images sequentially, updating the context set with newly generated images at each step. This corresponds to the following autoregressive model:
\begin{align}
  p\left(\left\{ \imtarget_k \right\}_k \ | \ \left\{ \anntarget_k \right\}_k, \ \context \right) &= \prod_k  p\left(\imtarget_k \ | \ \anntarget_k, \ \context_k \right) \\
 \context_k &=\contextset \cup \left\{ (\imtarget_p, \anntarget_p) \right\}_{p{<}k}.
\end{align}
Predictions for each target image will take into account all previously generated images, and can therefore capture the required dependencies. 

\paragraph{A unified task formulation} We apply our framework to four different general task specifications, which are distinguished by their target annotations $\anntarget$ and context set $\context$, as described in Table~\ref{tab:taskmodels}. 

\paragraph{Training objective} A range of modelling paradigms can be used to estimate the predictive distributions in Equation \ref{eq:preddist}, with DCTransformer with sparse DCT representations being a particular example. Other examples include conditional VAEs such as GQN~\cite{eslami2018neural}, conditional GANs as in pix2pix~\cite{pix2pix2017}, or conditional diffusion-based models~\cite{DBLP:conf/nips/HoJA20}. The kind of training objective depends on the model class used. In this work, we build on the DCTransformer, and therefore use the corresponding autoregressive training objective described in Section \ref{sec:background}.

\begin{figure}[t]
    \centering
    \begin{subfigure}[b]{0.475\linewidth}
        \includegraphics[width=\linewidth]{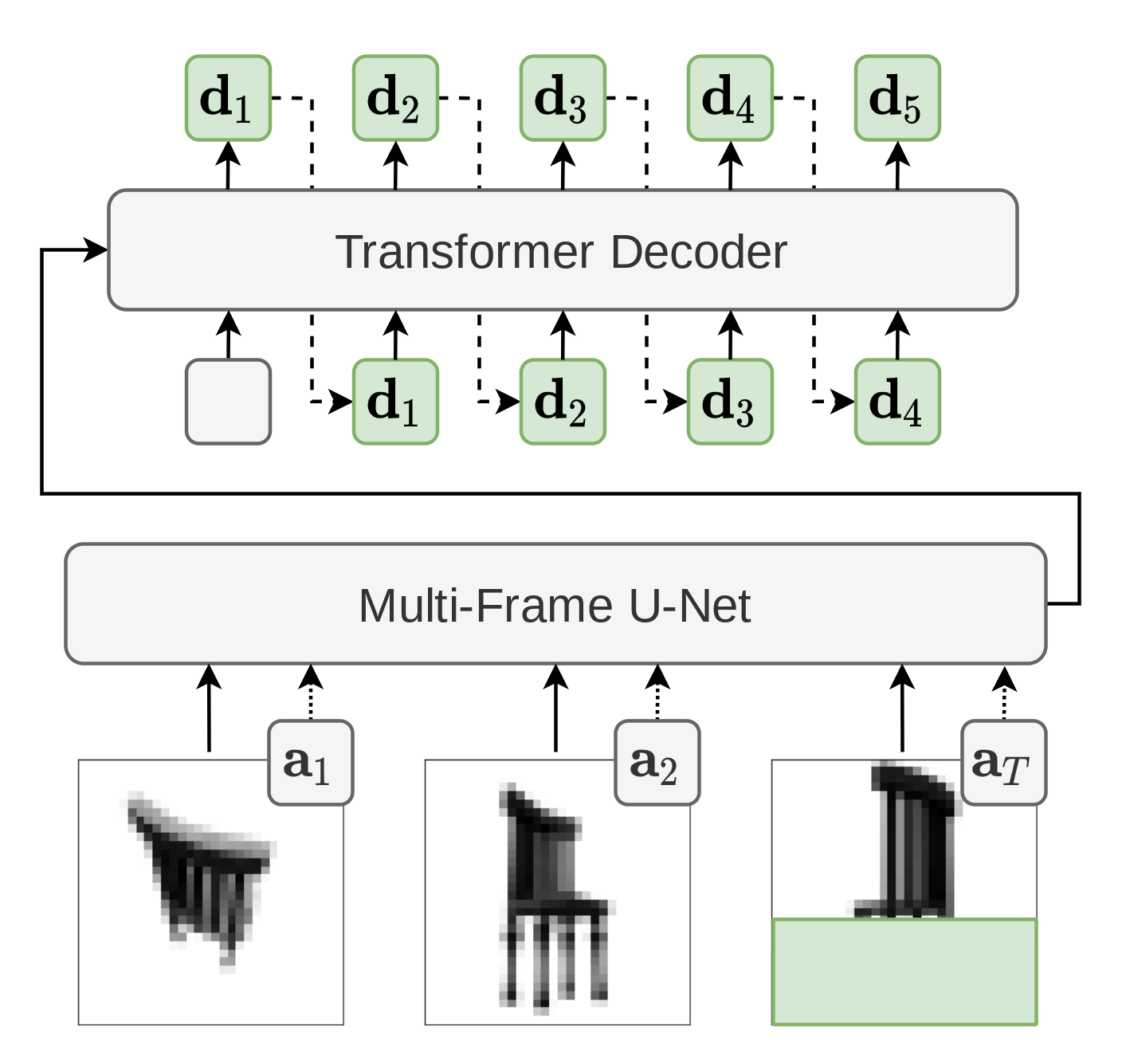}
        \caption{Transframer architecture.}
    \end{subfigure}
    \hfill
    \begin{subfigure}[b]{0.475\linewidth}
        \includegraphics[width=\linewidth]{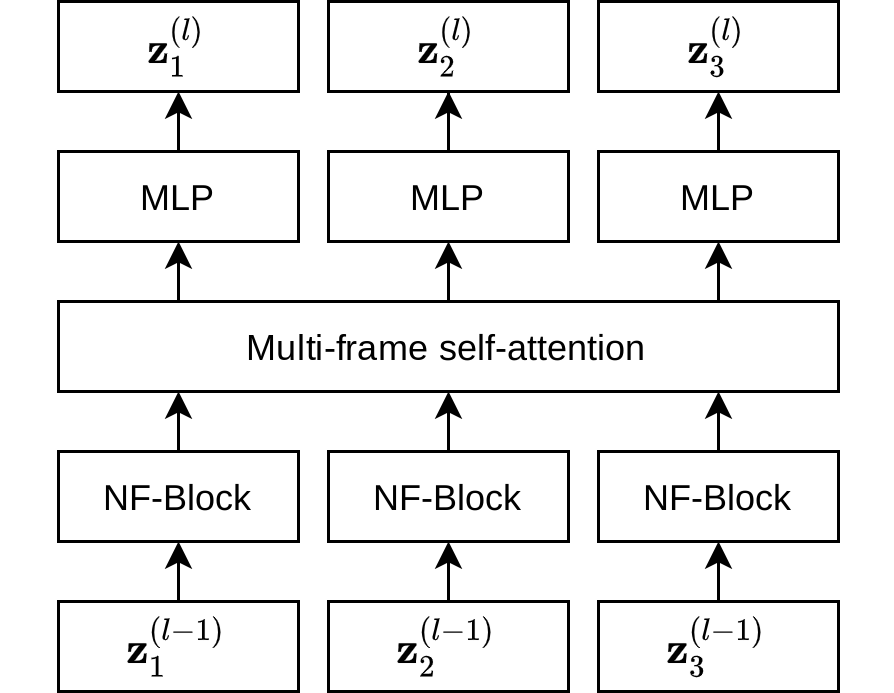}   
        \caption{Multi-frame U-Net block.}
    \end{subfigure}
    \caption{(a) \modelname~takes as input context DCT-images (left and middle), as well as a partially observed DCT-image of the target (right) and additional annotations $\ann$. The inputs are processed by a multi-frame U-Net encoder, which operates at a number of spatial resolutions. U-Net outputs are passed to a DCTransformer decoder via cross-attention, which autoregressively generates a sequence of DCT tokens corresponding to the unseen portion of the target image (shown in green). (b) Multi-frame U-Net blocks consist of NF-Net convolutional blocks, multi-frame self-attention blocks - which exchange information across input frames - and a Transformer-style residual MLP.}
    \label{fig:arch}
\end{figure}

\subsection{Transframer architecture}
\label{subsec:architecture}

To estimate predictive distributions over target images, we require an expressive generative model that can produce diverse, high-quality outputs. DCTransformer (Section \ref{sec:background}, \cite{nash2021generating}) produces compelling results on single image domains, but is not designed to condition on the multi-image context sets $\contextset$ that we require. As such, we extend DCTransformer (Section \ref{sec:background}, \cite{nash2021generating}) to enable image and annotation-conditional predictions. In particular, we replace DCTransformer's Vision-Transformer-style encoder, which operates on single (partially observed) DCT images, with a multi-frame U-Net architecture, that processes a set of annotated context frames along with the partially observed target frame. As in DCTransformer, we pass output embeddings from the encoder to a Transformer-based decoder, and autoregressively predict sparse DCT tokens (Figure \ref{fig:arch}a).

\paragraph{Multi-frame U-Net} The input to the U-Net is a sequence consisting of $N$ context DCT frames (See Section \ref{sec:background}), and a partially observed target DCT frame. Annotation information is provided in the form of vectors associated with each input frame. Initially, the DCT image channels are linearly projected to an embedding dimensionality $E$ using $1\times1$ convolutions. We add horizontal and vertical positional embeddings to the embedded DCT images, and incorporate annotation information $\ann$ by adding linearly projected annotation vectors. 

The core component of our U-Net is a computational block that first applies a shared NF-ResNet convolutional block \cite{DBLP:conf/icml/BrockDSS21} to each input frame, and then applies a Transformer-style self-attention block to aggregate information across frames (Figure \ref{fig:arch}b). NF-ResNet blocks consist of grouped convolutions followed by a squeeze-and-excite layer \cite{DBLP:conf/cvpr/HuSS18}, and are designed for efficient performance on TPUs~\cite{tpu}. The self-attention block is based on a standard Transformer block~\cite{vaswani2017attention}, with a self-attention layer applied before an MLP. We use one of two self-attention types: In the first case, self-attention is applied to corresponding spatial locations across the frame feature maps. In the latter case, self-attention is applied both spatially and across frames. In some cases, we omit self-attention entirely to reduce memory consumption. We use multi-query attention~\cite{shazeer2019}, rather than the standard multi-head attention in order to improve sampling speed.

As in a standard U-Net, our multi-frame variant has a downsampling stream, where feature maps are spatially downsampled for more efficient processing, and an upsampling stream where feature maps are spatially upsampled back to the input resolution. Skip-connections are used to propagate information from the downsampling stream to the upsampling stream directly, bypassing low-resolution information bottlenecks. In both streams, multiple convolutional blocks are applied at a given resolution before an upsampling or downsampling block changes the spatial resolution. We use a symmetrical structure with the same number of block at each resolution in both streams.

\subsection{Residual DCT representations}
\label{subsec:residuals}

\bgroup
\def\arraystretch{0.25}
\begin{figure}[t]
    \centering
    \begin{subfigure}[b]{0.39\linewidth}
        \centering
        \setlength\tabcolsep{0cm}
        \begin{tabular}{>{\centering\arraybackslash} m{0.2\linewidth}m{0.8\linewidth}}
        \multicolumn{2}{c}{\includegraphics[width=\linewidth]{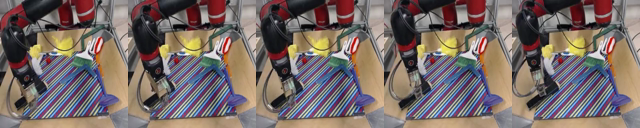}} \\
        \includegraphics[width=\linewidth]{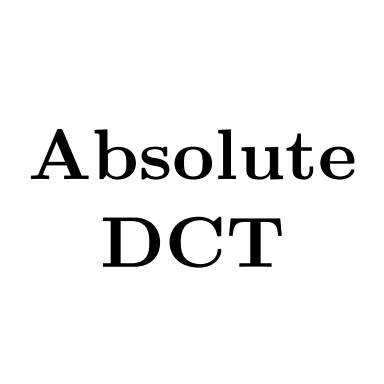} & \includegraphics[width=\linewidth]{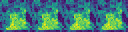} \\
        \includegraphics[width=\linewidth]{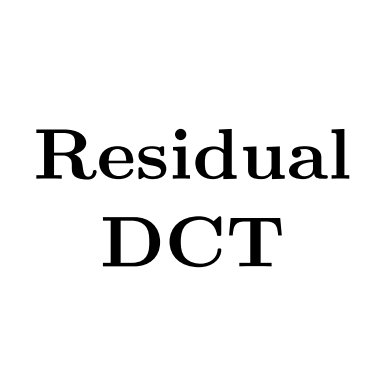} &\includegraphics[width=\linewidth]{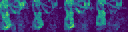} \\
        \multicolumn{2}{c}{\includegraphics[width=\linewidth]{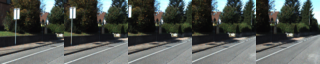}}\\
        \includegraphics[width=\linewidth]{graphics/abs_text.png} & \includegraphics[width=\linewidth]{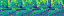}\\
        \includegraphics[width=\linewidth]{graphics/resid_text.png} &\includegraphics[width=\linewidth]{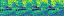} 
        \end{tabular}
        \caption{Sparsity heatmaps}
    \end{subfigure}\hfill
    \begin{subfigure}[b]{0.59\linewidth}
        \centering
        \begin{subfigure}[b]{\linewidth}
            \centering
            \includegraphics[width=\linewidth]{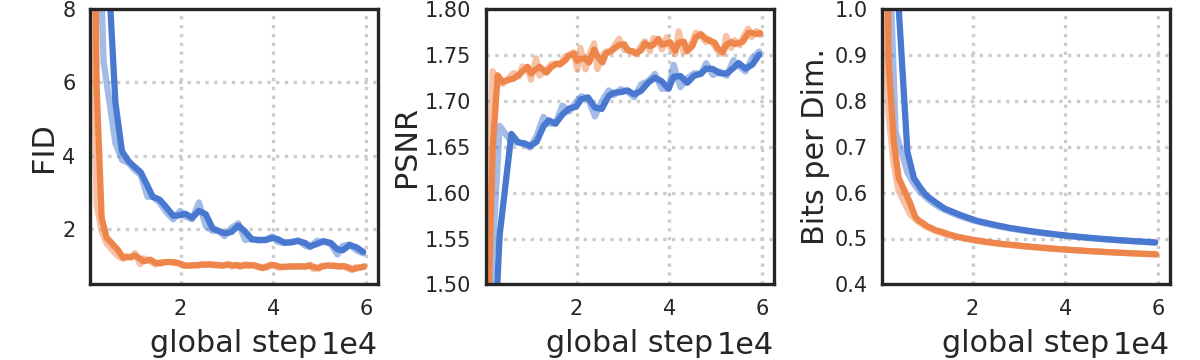}
            \caption{BAIR (64x64)}
        \end{subfigure}
        \begin{subfigure}[b]{\linewidth}
            \centering
            \includegraphics[width=\linewidth]{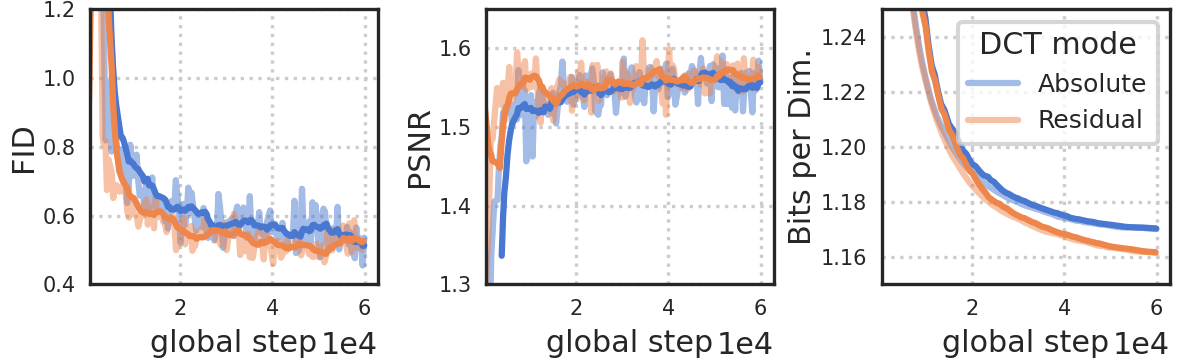}
            \caption{KITTI (64x64)}
        \end{subfigure}
    \end{subfigure}
    \caption{(a) Comparing the sparsity of absolute and residual DCT representations for RoboNet (128x128) and KITTI videos. RoboNet consists of static videos with only a few moving elements, so residual frame representations yield a substantial increase in sparsity. KITTI videos typically feature a moving camera, resulting in differences almost everywhere in consecutive frames. In this case the sparsity benefits of residuals is minor. (b) Evaluation metrics for absolute and residual representations on BAIR and KITTI.}
    \label{fig:residuals}
\end{figure}
\egroup

Consecutive video frames often exhibit substantial temporal redundancy. As an extreme case, in videos with static cameras and backgrounds, pixel data may be almost entirely unchanged across adjacent frames. In the DCTransformer data format (Section~\ref{sec:background}), a sparse image representation is obtained by storing the positions and values of non-zero DCT co-efficients. When working with video data, we can potentially increase the degree of sparsity by storing DCT data only where it changes relative to the previous frame. 

Figure \ref{fig:residuals} visualizes the sparsity of absolute and residual DCT representations for two datasets. The sparsity gains of the residual representation are significant in the static-camera RoboNet dataset~\cite{dasari2019robonet}. However, for the KITTI dataset~\cite{Geiger2012CVPR}, which features a continuously moving camera, the sparsity gains are diminished. In our experiments, we use residual DCTs for video prediction, and absolute DCTs otherwise.

\section{Experiments}
\label{sec:results}
To illustrate the flexibility and universality of our framework, and to test the  performance of \modelname, we use a range of datasets and tasks, covering predictive video modelling, few-shot novel view synthesis, and a number of classic computer vision tasks. Our focus is on generality, and the ease of expressing a given task in the presented framework. Please see the appendix for model hyperparameters, training details, and additional ablation studies. A selection of example output videos can be seen at \url{https://sites.google.com/view/transframer}.

\subsection{Video modelling}
\label{subsec:video_experiments}
We train \modelname to predict the next frame given a series of input video frames. We supply one-hot time-steps as annotations as described in Table \ref{tab:taskmodels} (row 1), where time is specified relative to the target frame. For each video modelling task, we specify a minimum and maximum number of context frames. During training, for each element of a batch, we uniformly at random choose the number of context frames in the specified range, and pad the sequence to the maximum number of frames. In general, we choose the minimum number of frames to be equal to the number of context frames provided at test time.

\begin{table}[t]
    \centering
    \caption{Fréchet video distance for generated videos on BAIR and Kinetics600 datasets. For BAIR, we use a single context frame and generate 15 frames. For Kinetics600, we use 5 context frames and generate 11 frames.}
    \begin{tabular}{lr}\toprule
        \textbf{BAIR (64x64)} &FVD$\downarrow$ \\\midrule
        DVD-GAN-FP~\cite{clark2019adversarial} &109.8 \\
        VideoGPT~\cite{DBLP:journals/corr/abs-2104-10157} &103.3 \\
        TrIVD-GAN-FP~\cite{DBLP:journals/corr/abs-2003-04035} &103.3 \\
        \modelname (Ours) &100.0 \\
        Video Transformer~\cite{DBLP:conf/iclr/WeissenbornTU20} &94.0 \\
        FitVid~\cite{DBLP:journals/corr/abs-2106-13195} & \textbf{93.6} \\\bottomrule
    \end{tabular}\qquad
    \begin{tabular}{lr}\toprule
        \textbf{Kinetics600 (64x64)} &FVD$\downarrow$ \\\midrule
        Video Transformer~\cite{DBLP:conf/iclr/WeissenbornTU20} &170.0 \\
        DVD-GAN-FP~\cite{clark2019adversarial} &69.1 \\
        Video VQ-VAE~\cite{DBLP:journals/corr/abs-2103-01950} &64.3 \\
        TrIVD-GAN-FP~\cite{DBLP:journals/corr/abs-2003-04035} & 27.7\\
        \modelname (Ours) &\textbf{25.4}
        \\\bottomrule
    \end{tabular}
    \label{tab:fvd_kinetics_bair}
\end{table}

\begin{table}[t]
    \centering
    \caption{Sample metrics for generated videos on KITTI and action-conditional RoboNet datasets. For KITTI, we use 5 context frames and generate 25 frames. For RoboNet, we use 2 context frames and generate 10 frames.}
    \begin{tabular}{lrrrrrrrr}\toprule
        &\multicolumn{4}{c}{\textbf{KITTI (64x64)}} &\multicolumn{4}{c}{\textbf{RoboNet (64x64)}} \\\cmidrule(lr){2-5}\cmidrule(lr){6-9}
        &FVD$\downarrow$ &PSNR$\uparrow$ &SSIM$\uparrow$ &LPIPS$\downarrow$ &FVD$\downarrow$ &PSNR$\uparrow$ &SSIM$\uparrow$ &LPIPS$\downarrow$ \\\midrule
        SVG~\cite{DBLP:conf/nips/VillegasPKELL19} &1217.3 &15.0 &41.9 &0.327 &123.2 &23.9 &87.8 &0.060\\
            GHVAE~\cite{DBLP:conf/cvpr/WuNM0F21} &552.9 &15.8 &51.2 &0.286 &95.2 &24.7 &89.1 &0.036\\
        FitVid~\cite{DBLP:journals/corr/abs-2106-13195} &884.5 &17.1 &49.1 &0.217 &62.5 &28.2 &89.3 &0.024\\ 
        \modelname (Ours) &\textbf{260.4} &\textbf{17.9} &\textbf{54.0} &\textbf{0.112} &\textbf{53.0} &\textbf{31.3} &\textbf{94.1} &\textbf{0.013}\\
        \bottomrule
    \end{tabular}
    \label{tab:metrics_robonet_64}
\end{table}

\paragraph{Metrics}  
For video generation we use the following metrics: Structural Similarity Index Measure (SSIM,~\cite{DBLP:journals/tip/WangBSS04}), Peak Signal-to-noise Ratio (PSNR, \cite{huynh2008scope}), Learned Perceptual Image Patch Similarity (LPIPS, \cite{DBLP:conf/cvpr/ZhangIESW18}) and Fréchet Video Distance (FVD, \cite{DBLP:journals/corr/abs-1812-01717}). FVD measures distributional similarity between sets of ground truth and generated videos, and is sensitive to image quality and temporal coherence. For RoboNet and KITTI we follow the literature and report the best SSIM, PSNR, and LPIPS scores over 100 trials for each video. For consistency with previous work we only report test-set FVD for Kinetics600 and BAIR.

\paragraph{Datasets} 
We first evaluate our model on \textbf{BAIR}~\cite{bair}, which is one of the most well studied video modelling datasets, consisting of short video clips of a single robot arm interactions in the unconditional setting. Table~\ref{tab:fvd_kinetics_bair} shows the FVD score obtained by our model. While our model improves on some strong baselines such as TrIVD-GAN, it is out-performed by FitVid and Video Transformer. However, BAIR's small test set consisting of only 256 samples makes precise comparisons challenging, as discussed by Luc et al.\ ~\cite{DBLP:journals/corr/abs-2003-04035}. Moreover, both methods under-perform \modelname in all other benchmarks, with larger test sets. 

\textbf{Kinetics600}~\cite{DBLP:journals/corr/abs-1808-01340} is an action recognition dataset, consisting of video clips of dynamic human actions across 600 activities, such as sailing, chopping, and dancing. It is a challenging dataset for video modelling, as the scenes feature complex motion, as well as sudden scene transitions. We use the same experimental set-up as DVD-GAN~\cite{clark2019adversarial}, operating at $64{\times}64$ resolution, and generate 11 frames given 5 context frames. We report test-set FVD, for 50k frames, sampling the context frames uniformly at random from the test-set clips. Our model obtains an FVD of 25.4, improving over the previous state-of-the-art model TrIVD-GAN-FP~\cite{DBLP:journals/corr/abs-2003-04035}.

The \textbf{KITTI} dataset~\cite{Geiger2012CVPR} contains long video clips of roads and surrounding areas taken from the viewpoint of a car driver. It is less challenging than Kinetics600 due to smoother camera motion, and a more constrained visual setting (roads, cars, etc.). However it is a relatively small dataset, which makes generalization challenging. As such, we employ the augmentation methods of Babaeizadeh et al.\ ~\cite{DBLP:journals/corr/abs-2106-13195} to reduce overfitting. We follow the evaluation regime of Villegas et al.\ ~\cite{DBLP:conf/icml/VillegasYZSLL17}, operating at $64{\times}64$ resolution, and generate 25 frames given 5 context frames, with test clips taken from 3 longer test clips at 5-frame strides. Table~\ref{tab:metrics_robonet_64} shows that our model's performance improves on comparable methods on all metrics, with the improvement in LPIPS and FVD being the most substantial.

\newcommand{\imwidthplot}{2.6cm}
\begin{figure}[t]
    \centering
    \resizebox{\linewidth}{!}{
        \begin{tabular}{ccccc} 
          Context ($t{=}0$) & Context ($t{=}1$) & Prediction ($t{=}6$) & Actual ($t{=}6$) & Errors ($t{=}6$) \\
          \includegraphics[width=\imwidthplot]{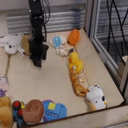} & \includegraphics[width=\imwidthplot]{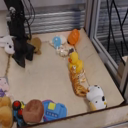} & 
          \includegraphics[width=\imwidthplot]{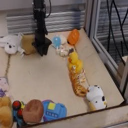} & \includegraphics[width=\imwidthplot]{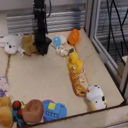} & \includegraphics[width=\imwidthplot]{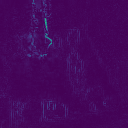}\\
          & & Prediction ($t{=}11$) & Target ($t{=}11$) & Errors ($t{=}11$)\\
          & & \includegraphics[width=\imwidthplot]{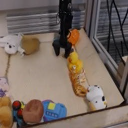} & \includegraphics[width=\imwidthplot]{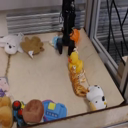} & \includegraphics[width=\imwidthplot]{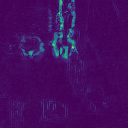}\\
        \end{tabular}
    }
    \caption{Action-conditional video generation on the RoboNet test set at 128x128 resolution. The action sequence for the output frames is provided to the model, and the frames are autoregressively generated, with a context length of up to 4.}
    \label{fig:robonet_diffs}
\end{figure}

To evaluate our model in the action-conditional setting, we use \textbf{RoboNet}~\cite{dasari2019robonet}, which consists of short video clips of robot arms interacting with objects, and provides robot action annotations as 5-dimensional vectors. The actions relate to the robot end-effector and gripper joint. We train at 64x64 and 128x128 resolutions, and evaluate using 2 context frames and 10 sampled frames on the test set specified by FitVid~\cite{DBLP:journals/corr/abs-2106-13195}. To process the input actions, we linearly project the 5-dimensional action-vectors and add them to input DCT image embeddings in the encoder. Table~\ref{tab:metrics_robonet_64} shows the model's performance in comparison to existing methods, and Figure~\ref{fig:robonet_diffs} shows an example of our model's outputs at 128x128 resolution. At 64x64 resolution, our model improves on alternatives in every metric by a large margin. At 128x128, we could not find comparable previous work, so we report our results in Appendix E for future comparisons. 
\renewcommand{\imwidthplot}{1.25cm}
\begin{figure}[t]
\centering
\begin{tabular}{cccccccc} 
      Context  & 1s & 2s & 4s & 8s & 16s & 30s \\
     \hline\\
     \includegraphics[width=\imwidthplot]{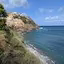} & 
    %  \textcolor{context}{\fbox{\includegraphics[width=\imwidthplot]{graphics/30s_video/multi_image/sardinia_1/output_0001.png}}} & 
     \includegraphics[width=\imwidthplot]{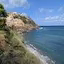} &
     \includegraphics[width=\imwidthplot]{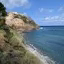} &
    \includegraphics[width=\imwidthplot]{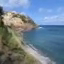} & 
    \includegraphics[width=\imwidthplot]{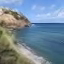} & 
    \includegraphics[width=\imwidthplot]{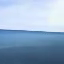} & 
    \includegraphics[width=\imwidthplot]{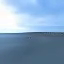}\\ 
    \includegraphics[width=\imwidthplot]{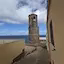} & 
    \includegraphics[width=\imwidthplot]{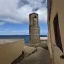} &
    \includegraphics[width=\imwidthplot]{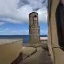} &
    \includegraphics[width=\imwidthplot]{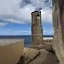} & 
    \includegraphics[width=\imwidthplot]{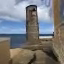} & 
    \includegraphics[width=\imwidthplot]{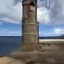} & 
    \includegraphics[width=\imwidthplot]{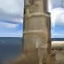}\\ 
    \includegraphics[width=\imwidthplot]{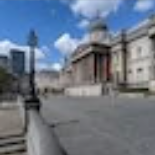} & 
    \includegraphics[width=\imwidthplot]{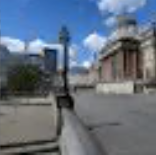} &
    \includegraphics[width=\imwidthplot]{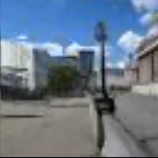} &
    \includegraphics[width=\imwidthplot]{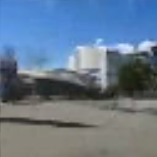} & 
    \includegraphics[width=\imwidthplot]{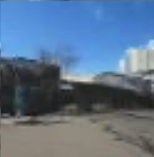} & 
    \includegraphics[width=\imwidthplot]{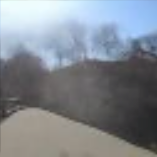} &
    \includegraphics[width=\imwidthplot]{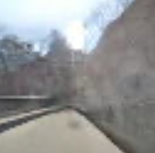}\\
    \includegraphics[width=\imwidthplot]{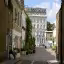} & 
    \includegraphics[width=\imwidthplot]{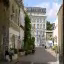} &
    \includegraphics[width=\imwidthplot]{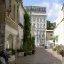} &
    \includegraphics[width=\imwidthplot]{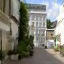} &
    \includegraphics[width=\imwidthplot]{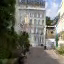} &
    \includegraphics[width=\imwidthplot]{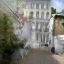} &
    \includegraphics[width=\imwidthplot]{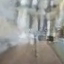}\\
\end{tabular}
    \caption{Given a single context frame we generate 30 seconds of video at 25fps. Full videos (including interpolated frames) and attributions for the context images are available at \url{https://sites.google.com/view/transframer}}
    \label{fig:30s_multi_image}
\end{figure}

\paragraph{Long range video generation} As an additional proof-of-concept, we push the temporal limits of our model and generate 30 second videos at 25 frames per second (fps) conditioned on a single input image. At this frame rate, videos consist of 750 frames total. Generating these frames sequentially is challenging for a number of reasons. First, our architecture would need to process a large number of frames to model long-range dependencies; scaling such a model would be computationally expensive. Second, simply unrolling a conventional model with only a few frames of context often leads to globally incoherent generations or collapse of the output as generation errors compound over time-steps. 

We address these issues with a two-stage procedure, where we first generate low fps videos using one model, and then interpolate to obtain the desired frame rate with another. We train the low fps model 1 fps videos with up to 15 context frames. At evaluation time we condition on a single context image, and sequentially predict 29 frames to yield a 1 fps video that is 30 seconds long in real time. We train a separate model using the same architecture to interpolate between consecutive 1 fps frames at 25 fps. During training, the interpolation model predicts a target frame conditioned on the penultimate frame as well as a single future frame. At test time the model takes consecutive 1fps frames as context, then generates the 25 fps frames successively. The first two rows in Table~\ref{tab:taskmodels} describe both stages, and Figure~\ref{fig:30s_multi_image} shows some example outputs.

\paragraph{Impact of residual DCT representations} As described in Section~\ref{subsec:residuals}, for video datasets where consecutive frames have a high level of temporal redundancy, images can be more compactly represented using the difference between DCT values in adjacent frames. We compare the data representations in Figure~\ref{fig:residuals} with training curves for residual and absolute DCT sequences for BAIR and KITTI. Our experiments on both datasets suggest that residual representations provide an advantage, which is greater on BAIR, likely due to the static backgrounds and associated sparsity.

\subsection{Novel view synthesis}
\label{subsec:view_synth_experiments}
We  train \modelname for novel view synthesis by conditioning on the target camera position, as well as collection of annotated input views. We supply camera views as context and target annotations as described in Table \ref{tab:taskmodels} (row 3), and sample uniformly a number of context views, up to a specified maximum.

\newcommand{\viewsynthidA}{0}
\newcommand{\viewsynthidB}{0}
\newcommand{\viewsynthidC}{10}
\newcommand{\viewsynthidD}{10}
\newcommand{\imheightplot}{0.83cm}
\newcommand{\viewtabletitlesize}{\scriptsize}
\begin{figure}[t]
    \centering
    \begin{subfigure}[t]{0.59\linewidth}
        \setlength\tabcolsep{0cm}
        \begin{tabular}{cc|c|c|cc|c} 
            \multicolumn{2}{c}{\viewtabletitlesize\underline{Context}} & \multicolumn{1}{c}{\viewtabletitlesize\underline{SRN}} & \multicolumn{1}{c}{\viewtabletitlesize\underline{PNeRF}} & \multicolumn{2}{c}{\viewtabletitlesize\underline{Ours}} & {\viewtabletitlesize\underline{Actual}} \\
            & \includegraphics[height=\imheightplot]{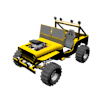} & \includegraphics[height=\imheightplot]{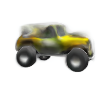} & \includegraphics[height=\imheightplot]{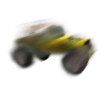} & \includegraphics[height=\imheightplot]{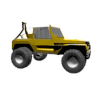} & \includegraphics[height=\imheightplot]{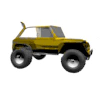} &\includegraphics[height=\imheightplot]{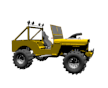} \\
            \includegraphics[height=\imheightplot]{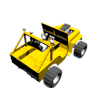} & \includegraphics[height=\imheightplot]{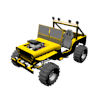} & \includegraphics[height=\imheightplot]{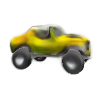} & \includegraphics[height=\imheightplot]{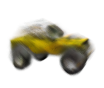} & \includegraphics[height=\imheightplot]{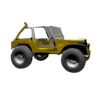} & \includegraphics[height=\imheightplot]{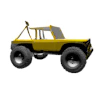} &\includegraphics[height=\imheightplot]{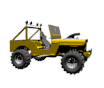}\\
            & \includegraphics[height=\imheightplot]{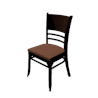} & \includegraphics[height=\imheightplot]{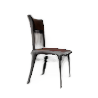} & \includegraphics[height=\imheightplot]{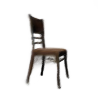} & \includegraphics[height=\imheightplot]{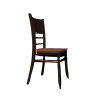} & \includegraphics[height=\imheightplot]{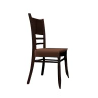} &\includegraphics[height=\imheightplot]{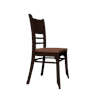} \\
            \includegraphics[height=\imheightplot]{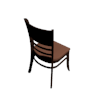} & \includegraphics[height=\imheightplot]{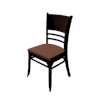} & \includegraphics[height=\imheightplot]{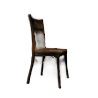} & \includegraphics[height=\imheightplot]{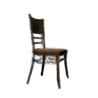} & \includegraphics[height=\imheightplot]{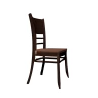} & \includegraphics[height=\imheightplot]{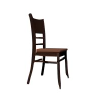} &\includegraphics[height=\imheightplot]{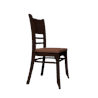}\\
        \end{tabular}
        \caption{ShapeNet}
    \end{subfigure}\hfill
    \begin{subfigure}[t]{0.39\linewidth}
        \begin{tabular}{cc|cc|c} 
            \multicolumn{2}{c}{\viewtabletitlesize\underline{Context}} &  \multicolumn{2}{c}{\viewtabletitlesize\underline{Ours}} & {\viewtabletitlesize\underline{Actual}} \\
            & \includegraphics[height=\imheightplot]{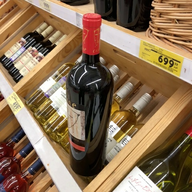} & \includegraphics[height=\imheightplot]{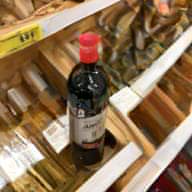} &  \includegraphics[height=\imheightplot]{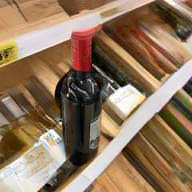} &\includegraphics[height=\imheightplot]{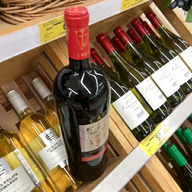} \\
            \includegraphics[height=\imheightplot]{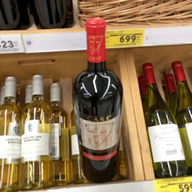} & \includegraphics[height=\imheightplot]{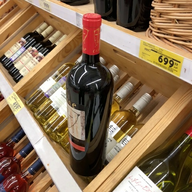} & \includegraphics[height=\imheightplot]{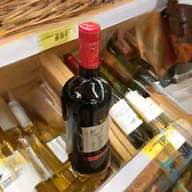} &  \includegraphics[height=\imheightplot]{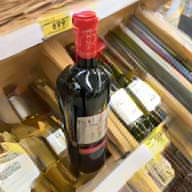} &\includegraphics[height=\imheightplot]{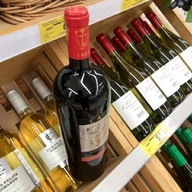} \\
            & \includegraphics[height=\imheightplot]{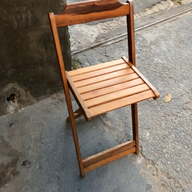} & \includegraphics[height=\imheightplot]{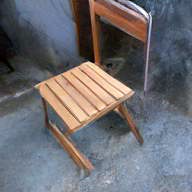} &  \includegraphics[height=\imheightplot]{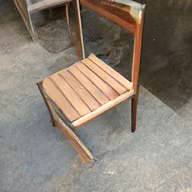} &\includegraphics[height=\imheightplot]{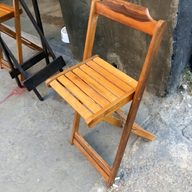} \\
            \includegraphics[height=\imheightplot]{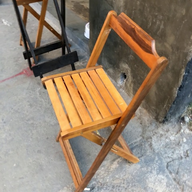} & \includegraphics[height=\imheightplot]{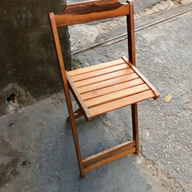} & \includegraphics[height=\imheightplot]{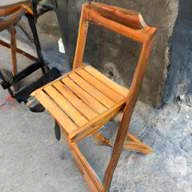} &  \includegraphics[height=\imheightplot]{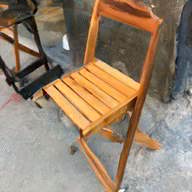} &\includegraphics[height=\imheightplot]{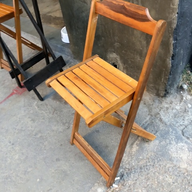} \\
        \end{tabular}
        \caption{Objectron}
    \end{subfigure}
    \caption{View synthesis on ShapeNet (a) and Objectron (b).
    For ShapeNet we also show the corresponding outputs from SRN~\cite{sitzmann2019scene} and PixelNeRF~\cite{DBLP:conf/cvpr/YuYTK21}.}
    \label{fig:view_syn}
\end{figure}

\begin{table}[t]
    \centering
    \caption{View synthesis results on ShapeNet.}
    \begin{tabular}{lrrrrrrrr}\toprule
        & \multicolumn{4}{c}{\textbf{Chair (1 view)}} & \multicolumn{4}{c}{\textbf{Chair (2 view)}} \\ 
        \cmidrule(lr){2-5}\cmidrule(lr){6-9}
        &FID$\downarrow$ &PSNR$\uparrow$ &SSIM$\uparrow$ &LPIPS$\downarrow$ &FID$\downarrow$ &PSNR$\uparrow$ &SSIM$\uparrow$ &LPIPS$\downarrow$ \\\midrule
        PixelNeRF & 132.86 &\textbf{23.72} &\textbf{0.91} & 0.13 & 101.22 &\textbf{26.20} &\textbf{0.94} & 0.08\\
        \modelname (Ours) & \textbf{51.09} & 22.85 & \textbf{0.91} & \textbf{0.07} & \textbf{46.17}& 23.77 & 0.92 & \textbf{0.06} \\\\
        & \multicolumn{4}{c}{\textbf{Car (1 view)}} & \multicolumn{4}{c}{\textbf{Car (2 view)}} \\ 
        \cmidrule(lr){2-5}\cmidrule(lr){6-9}
        &FID$\downarrow$ &PSNR$\uparrow$ &SSIM$\uparrow$ &LPIPS$\downarrow$ &FID$\downarrow$ &PSNR$\uparrow$ &SSIM$\uparrow$ &LPIPS$\downarrow$ \\\midrule
        SRN & 142.85 &22.25 & 0.89 & 0.13 & 133.42  &24.84 &0.92 & 0.11\\
        PixelNeRF & 161.51 &\textbf{23.17} &\textbf{0.90} & 0.15 & 129.65 &\textbf{25.66} &\textbf{0.94} & 0.10\\
        \modelname (Ours) & \textbf{74.94} & 21.94 & 0.89 & \textbf{0.10} & \textbf{62.70}& 23.34 & 0.91 & \textbf{0.08}\\
        \bottomrule
    \end{tabular}
    \label{tab:shapenet}
\end{table}

\paragraph{Metrics}  
We follow Yu et al.\ ~\cite{DBLP:conf/cvpr/YuYTK21} in using PSNR and SSIM to evaluate our model's few-shot view synthesis performance. These metrics are informative, but problematic in the case where the target view is not well determined by the input views. In these cases, the single prediction that optimizes PSNR will be a blurred average over possible target views. This penalizes models such as ours that make probabilistic predictions, rather than a single point estimate. Hence, we introduce LPIPS and Fréchet Inception Distance (FID, \cite{DBLP:conf/nips/HeuselRUNH17}) as additional metrics, both of which correlate with human perceptual preferences to a greater degree than the alternatives. We calculate per-scene FID between two sets: The ground truth set of the combined input and target views, and the sampled set consisting of all generated views combined with the input views.

\paragraph{Datasets} We evaluate our model on \textbf{ShapeNet} benchmarks, in particular the chair and car subsets used by Yu et al.\ ~\cite{DBLP:conf/cvpr/YuYTK21}. The dataset consists of renders of 3D objects from the ShapeNet database. We train a single model across both classes, and don't provide class labels as input. As in Yu et al.\ ~\cite{DBLP:conf/cvpr/YuYTK21}, we evaluate using either 1 or 2 context views, and predict the remaining views in a 251-frame test-set. To process the input views we flatten and linearly project the 9-dimensional camera matrix and add them to input DCT image embeddings in the encoder. Table~\ref{tab:shapenet} compares our model to alternative approaches, and Figure~\ref{fig:view_syn} shows example results. Our model achieves worse PSNR and SSIM scores than baseline methods, but performs better on LPIPS and FID. As mentioned earlier, our model outputs samples, rather than a point estimate. As such it is unsurprising that it is outperformed on PSNR and SSIM by methods that output point estimates. Qualitatively, we observe that our model produces sharper outputs that capture the input views' textures, compared to somewhat blurry outputs for unseen views for PixelNeRF and SRN. 

As a more realistic view-synthesis task, we train and evaluate on the \textbf{Objectron} dataset. Objectron consists of short, object-centered clips, with full object and camera pose annotations. The dataset features a diverse array of everyday objects, such as bottles, chairs and bikes, recorded in a variety of settings. To the best of our knowledge, no prior works have trained view-synthesis models in the few-shot setting on this dataset. We compare the performance of our model for varying numbers of context views, and describe our experimental setup in detail in Appendices F and G for future reference.  Figure~\ref{fig:view_syn} shows example outputs for two scenes. When given a single input view, the model produces coherent outputs, but misses some features such as the crossed chair legs. Given two views, these ambiguities are resolved to a large extent.

\subsection{Multi-task Vision}
\label{subsec:multi_task_experiments}

\renewcommand{\imwidthplot}{1.6cm}
\renewcommand{\viewtabletitlesize}{\small}
\newcommand{\vertlabel}[1]{\rotatebox{90}{\scriptsize \parbox{1cm}{#1}}}
\begin{figure}[t]
    \centering
    \resizebox{\linewidth}{!}{
        \setlength\tabcolsep{0.075cm}
        \begin{tabular}{cc|cc|ccc|cc|c}
            & \multicolumn{1}{c}{\viewtabletitlesize\underline{Context}} &  \multicolumn{2}{c}{\viewtabletitlesize\underline{Predictions}} & {\viewtabletitlesize\underline{Actual}} & &  \multicolumn{1}{c}{\viewtabletitlesize\underline{Context}} &  \multicolumn{2}{c}{\viewtabletitlesize\underline{Predictions}} & {\viewtabletitlesize\underline{Actual}} \\
              \vertlabel{Semantic Segm. \\ (Cityscapes)} & \includegraphics[width=\imwidthplot]{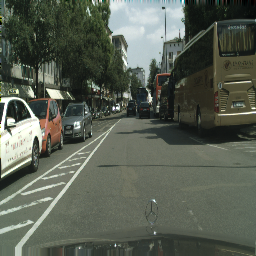} & \includegraphics[width=\imwidthplot]{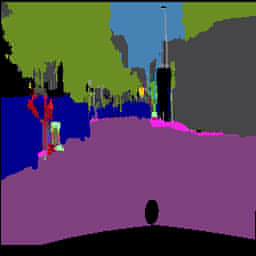} &  \includegraphics[width=\imwidthplot]{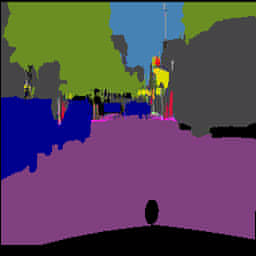} & \includegraphics[width=\imwidthplot]{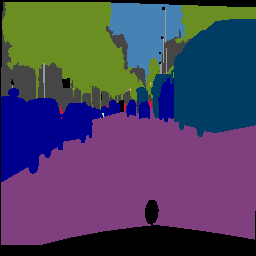} & 
              \vertlabel{Classification \\ (ImageNet)} & \includegraphics[width=\imwidthplot]{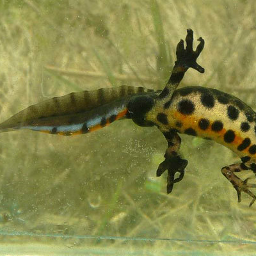} &
             \includegraphics[width=\imwidthplot]{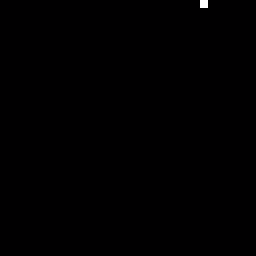} & \includegraphics[width=\imwidthplot]{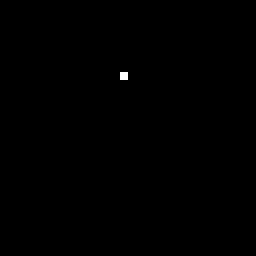} & \includegraphics[width=\imwidthplot]{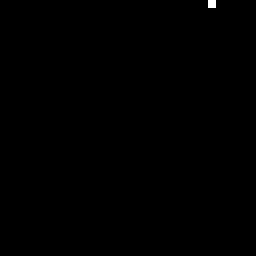} \\
              \vertlabel{Semantic Segm. \\ (PASCAL)} & \includegraphics[width=\imwidthplot]{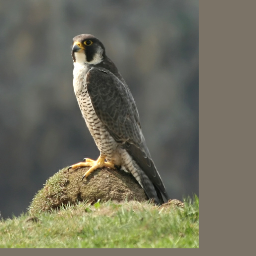} & \includegraphics[width=\imwidthplot]{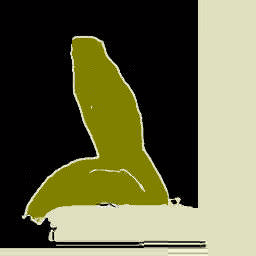} & \includegraphics[width=\imwidthplot]{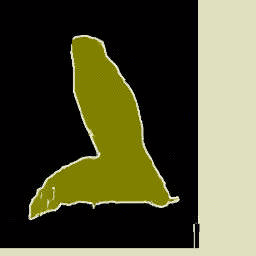} &  \includegraphics[width=\imwidthplot]{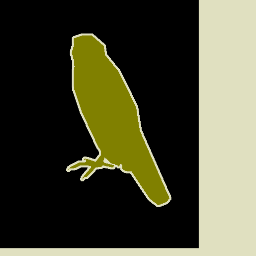} & 
              \vertlabel{Flow prediction \\ (K600)} & \includegraphics[width=\imwidthplot]{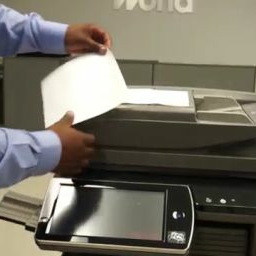} &
             \includegraphics[width=\imwidthplot]{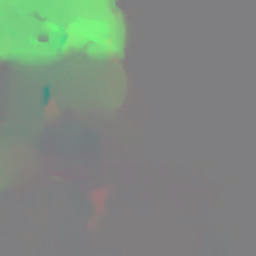} & \includegraphics[width=\imwidthplot]{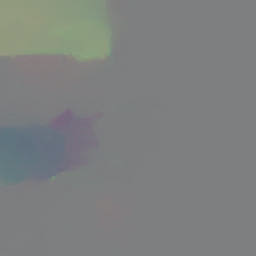} & \includegraphics[width=\imwidthplot]{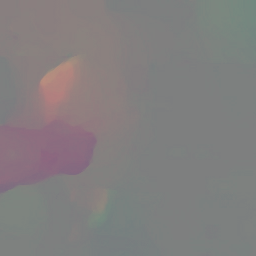} \\ \vertlabel{Detection \\ (COCO)} &  \includegraphics[width=\imwidthplot]{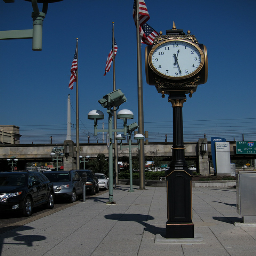}  &
              \includegraphics[width=\imwidthplot]{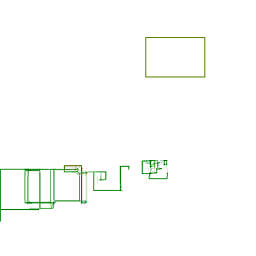} & \includegraphics[width=\imwidthplot]{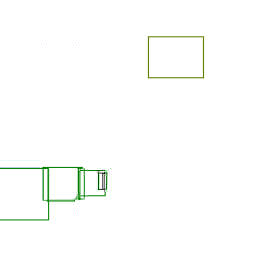} &  \includegraphics[width=\imwidthplot]{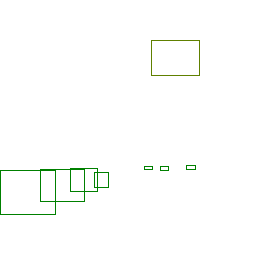} & 
              \vertlabel{Video modelling \\ (YT8M)} & \includegraphics[width=\imwidthplot]{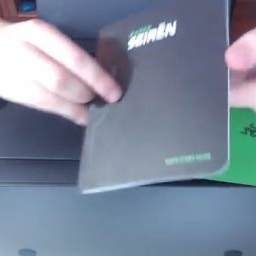} &
             \includegraphics[width=\imwidthplot]{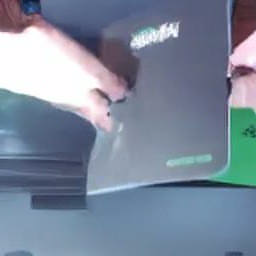} & \includegraphics[width=\imwidthplot]{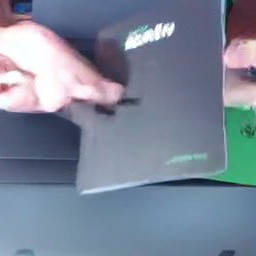} & \includegraphics[width=\imwidthplot]{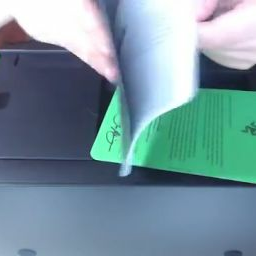}\\ \vertlabel{Scene parsing \\ (COCO)} & \includegraphics[width=\imwidthplot]{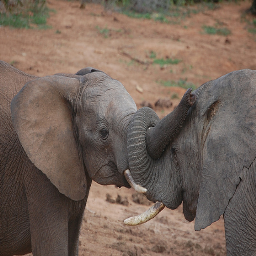} &
             \includegraphics[width=\imwidthplot]{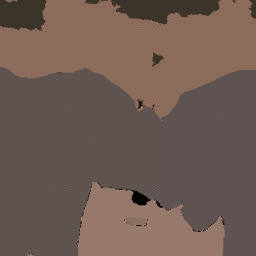} & \includegraphics[width=\imwidthplot]{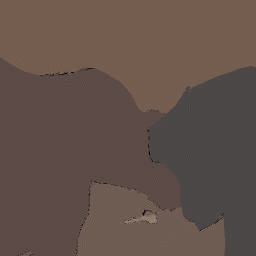} & \includegraphics[width=\imwidthplot]{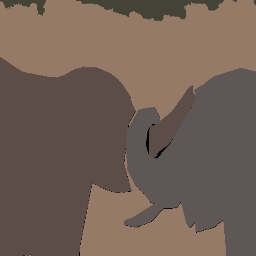} & 
             \vertlabel{Depth \\ estimation \\ (NYUDepth)} & \includegraphics[width=\imwidthplot]{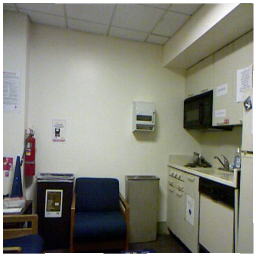} &
             \includegraphics[width=\imwidthplot]{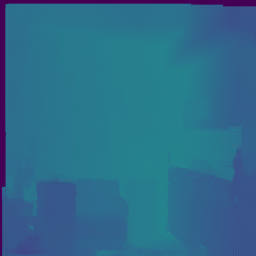} & \includegraphics[width=\imwidthplot]{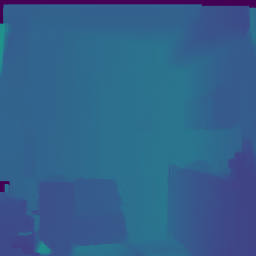} & \includegraphics[width=\imwidthplot]{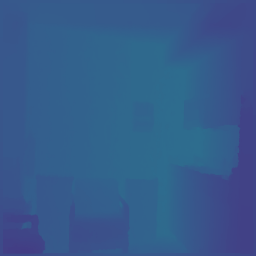} \\
        \end{tabular}
        }
    \caption{Example samples on the validation set of our multi-task setup, obtained using a single model. The model has no task specific parameters; in all cases it makes predictions conditioned on a context frame and a task embedding vector.}
    \label{fig:multitask}
\end{figure}
Different computer vision tasks are commonly handled using specialized (often intricately so) architectures and loss functions. In this section, we report an experiment where we jointly trained a single \modelname  model using the same loss function on 8 different tasks and datasets:  optical flow prediction from a single image, object classification, detection and segmentation, semantic segmentation (on two datasets), future frame prediction and depth estimation. We reduce all tasks to the same next-frame prediction task and do not adapt losses or architectural components. For the classification task, we use the ImageNet dataset~\cite{deng2009imagenet} and consider predicting \emph{one-hot images}, where each class is encoded as a 8x8 white patch on a black background. As image annotations we use a one-hot vector that determines the desired type of outputs, e.g., depth or optical flow. In all cases we used 256x256 resolution frames.

We trained a model using batches composed of randomly sampled elements from each dataset, rebalancing them slightly to accommodate their varied sizes. Several qualitative samples from the validation set are shown in Figure~\ref{fig:multitask}. It shows that the model learns to generate diverse samples across completely different tasks. On some tasks, such as Cityscapes, the model produces qualitatively good outputs, but model outputs on tasks like future frame prediction and bounding box detection are of variable quality, suggesting these are more challenging to model in this setting. Full experimental details can be found in Appendix H, along with quantitative performance metrics for the ImageNet classification and CityScapes semantic segmentation.

\section{Conclusion}
\label{sec:conclusion}
We introduced a general-purpose framework for conditional image prediction which unifies a range of image modelling and vision tasks. Our architecture, \modelname, models the conditional probability of image frames given target annotations and one or more context frames. Our results on video modelling and novel view synthesis confirm the quantitative strength of our approach, while our experiments applying the model to 8 standard vision tasks, including depth estimation, object detection, and semantic segmentation, demonstrate its broad applicability. These results show that general-purpose probabilistic image models can support challenging image and video modelling tasks, as well as multi-task computer vision problems traditionally addressed with task-specific methods. 

\clearpage
\bibliographystyle{splncs04}
\bibliography{egbib}

%%%%%%%% APPENDIX

%%%%%%%%% BODY TEXT
\newpage
\appendix

\section{Other related work}
\label{sec:extra_related}
In this work we focus on generative video modelling (Sec.\ \ref{subsec:video_experiments}) and view synthesis (Sec.\ \ref{subsec:view_synth_experiments}) as the main applications of our proposed approach. In our multi-task experiments (Sec.\ \ref{subsec:multi_task_experiments}) we explore how our generative approach can be applied to classic computer vision tasks like object classification, detection, and semantic segmentation, without using task-specific objectives or architectures. Here we present some relevant related work in these areas.

\paragraph{Generative video modelling} Video modelling is the task of predicting a distribution over video sequences. It is typically a conditional generative modelling problem, where the context is often a number of prior video frames. A range of generative video modelling approaches have been developed, of which there are two main categories: 

The first treats video modelling as an autoregressive frame prediction problem, where each generated frame is added to the existing context and passed back in to the model until a sequence of frames is produced. Approaches in this class differ primarily on the class of generative model used to capture conditional frame distributions, with variational autoencoders (VAEs)~\cite{DBLP:journals/corr/abs-2106-13195,DBLP:conf/iclr/BabaeizadehFECL18}, generative adversarial networks (GANs)~\cite{DBLP:journals/corr/abs-1804-01523,DBLP:journals/corr/abs-2003-04035}, and autoregressive models~\cite{DBLP:conf/iclr/WeissenbornTU20,DBLP:journals/corr/abs-2104-10157,DBLP:journals/corr/abs-2103-01950,DBLP:conf/visapp/RakhimovVAZB21} being the most prominent examples. Our video modelling approach is an example of this class of methods, with the most closely related approaches being those that use autoregressive image models in combination with compressed data representations \cite{DBLP:conf/visapp/RakhimovVAZB21,DBLP:journals/corr/abs-2104-10157,DBLP:journals/corr/abs-2103-01950}.

The second category of video modelling approaches predicts entire sequences of video frames at once, optionally conditioned on additional context such as prior video frames. Among the more prominent approaches in this category include MoCoGAN~\cite{DBLP:conf/cvpr/Tulyakov0YK18}, and DVD-GAN~\cite{clark2019adversarial}, which both use adversarial approaches to generate videos in a single network pass. In the case of DVD-GAN, the authors found the model to perform poorly in the video completion setting relative to the unconditional setting, with successor model TRiVD-GAN~\cite{DBLP:journals/corr/abs-2003-04035} using the autoregressive frame generation approach described above to address this issue.

In terms of the long video generation, most prior works can generate very short (a second or two) videos. In parallel to our work,~\cite{ge2022long} propose a similar two-stage approach to long video generation where anchor generation and interpolation models are trained individually. Their work, however, does the interpolation directly in the latent space.

\paragraph{Few-shot view synthesis} Few-shot view synthesis involves generating images from novel viewpoints conditioned on a small number of context images from other viewpoints. This contrasts with the dense novel view synthesis popularized recently by NeRF~\cite{DBLP:conf/eccv/MildenhallSTBRN20}, where a large number of views ($100-300$) with good coverage of target viewpoints are provided as input.

One family of approaches use explicit knowledge about 3D rendering to inform neural function estimators~\cite{DBLP:conf/cvpr/YuYTK21,DBLP:conf/icml/DupontMCSSS20,DBLP:journals/corr/abs-2107-05775,DBLP:conf/icml/RematasMF21}. While the use of a 3D inductive bias aids generalization, most methods produce point estimates of unseen views, leading to blurring under uncertainty. This blurring effect is evident in the PixelNeRF, and SRN outputs shown in Figure \ref{fig:shapenet}. A second family of approaches are based on deep generative models. Of these, our model is most closely related to Generative Query Networks (GQN) and its extensions~\cite{eslami2018neural,DBLP:journals/corr/abs-1807-02033}, which output distributions over query views without the use of geometric priors. Other models including NeRF-VAE~\cite{DBLP:conf/icml/KosiorekSZMSMR21}, Epipolar GQN~\cite{DBLP:conf/nips/TobinZA19} and Generative Scene Networks~\cite{devries2021unconstrained} combine 3D geometric approaches with generative models to try and gain the benefits of both approaches.

\paragraph{Computer vision as generative image modelling} In Section \ref{subsec:multi_task_experiments} we show that a unified learning objective and architecture can tackle a wide range of computer vision tasks: semantic segmentation, object detection, optical flow estimation, depth estimation, etc.\ 
% can be treated as instances of a broader class of image to image model. 
We drew inspiration from \texttt{pix2pix}~\cite{pix2pix2017}, where general image-to-image translation tasks are tackled with a unified architecture and training objective. While most applications of \texttt{pix2pix} involve predicting RGB images (image colorization, completion, etc.\ ), the authors also demonstrated a small number of standard computer vision tasks including semantic segmentation. In this work we push this approach even further, treating object detection, and object classification as generative frame prediction, all in a multi-task setting. To our knowledge this work is the first example of these tasks being performed in this way.

Unlike \texttt{pix2pix}, our model outputs a distribution over target frames, which enables us to express uncertainty and sample multiple possible predictions. This is important when target frames are not fully determined by the inputs, for example when scene depth is uncertain due to projection ambiguities, or object segmentation labels are inconsistent across different annotators. Prior work on diverse predictions includes ensemble-based methods~\cite{DBLP:conf/nips/Lakshminarayanan17} and the use of multiple prediction heads in combination with specialized training objectives~\cite{DBLP:journals/corr/LeePCCB15,DBLP:conf/iccv/RupprechtLDB17,DBLP:journals/corr/abs-1802-07095}. Zhu et al.\ extend the \texttt{pix2pix} framework with latent variables to capture additional variability in target images, and address the mode-collapse issue using a loss that encourages an invertible latent-output mapping~\cite{DBLP:conf/nips/ZhuZPDEWS17}. Kohl et al.\ use a probabilistic U-Net to obtain probabilistic image segmentations, with a focus on medical imaging data~\cite{DBLP:conf/nips/KohlRMFLMERR18,DBLP:journals/corr/abs-1905-13077}. While this is similar in some ways to our approach, we go further: demonstrating not just semantic segmentation, but depth estimation, optical flow estimation, object detection and classification.

\section{Training details}
\label{sec:hyperparams}

For all video models excluding the interpolation model we used residual DCT representations (Section \ref{subsec:residuals}). For view-synthesis and multi-task vision models we used absolute DCT representations due to the lack of temporal redundancy between context and target frames in those domains. 

We train all models using Adam optimizer \cite{DBLP:journals/corr/KingmaB14}, using a cosine learning rate schedule that decays from 1e-3 to 1e-4 over the first 75\% of training steps after an initial linear warmup of 1000 steps. We use batch size 1024 for all models except for the multi-task model, which uses batch size 4096. We train on TPUv4 slices containing between 16 and 64 chips, using data-parallelism across chips and gradient accumulation steps to obtain the required batch size. 

Table \ref{tab:hyperparams} shows detailed hyperparameters for each domain. `Channel multipliers' refers to the number of U-Net channels at each resolution as a multiple of the base channels. `Residual blocks' refers to the number of blocks at each resolution. `Attention types' refers to the type of attention used at each resolution, where self-attention is applied on: the frame dimension (`F'), both frame and spatial dimensions (`SF'), or not at all (`N') as described in Section \ref{subsec:architecture}.

\begin{table*}[]
    \setlength\tabcolsep{0.5em}
    \centering
    \caption{Training, data and model settings used for each domain. RNet refers to RoboNet and K600 refers to Kinetics600. See sections \ref{subsec:architecture} and \ref{sec:hyperparams} for details on architectural components and additional training details.}
    \resizebox{\linewidth}{!}{
        \begin{tabular}{@{}lcccccccc@{}}
            \toprule
            \textbf{Data}                        & BAIR           & K600    & RNet-64 & RNet-128   & KITTI          & ShapeNet            & Objectron      & Multi-task     \\ \midrule
            Image resolution        & 64             & 64             & 64              & 128                 & 64             & 128                 & 192            & 256            \\
            DCT block size          & 4              & 4              & 4               & 4                   & 4              & 4                   & 8              & 8              \\
            DCT quality             & 99             & 85             & 95              & 95                  & 95             & 95                  & 65             & 72             \\
            Chroma downsampling     & No             & Yes            & No              & Yes                 & No             & Yes                 & Yes            & No             \\
            Min. context frames     & 1              & 5              & 2               & 2                   & 5              & 1                   & 1              & 1              \\
            Max. context frames     & 4              & 5              & 4               & 4                   & 5              & 3                   & 3              & 1              \\
                                    &                &                &                 &                     &                &                     &                &                \\
            \textbf{U-Net}                   &                &                &                 &                     &                &                     &                &                \\\midrule
            Channels                & 512            & 768            & 512             & 256                 & 512            & 256                 & 640            & 512            \\
            Channel multipliers     & {[}1, 2{]}     & {[}1, 2{]}     & {[}1, 2{]}      & {[}1, 2, 4{]}       & {[}1, 2{]}     & {[}1, 2, 4{]}       & {[}1, 2{]}     & {[}1, 2{]}     \\
            Residual blocks    & {[}3, 4{]}     & {[}2, 4{]}     & {[}3, 4{]}      & {[}2, 2, 3{]}       & {[}3, 4{]}     & {[}2, 2, 3{]}       & {[}2, 4{]}     & {[}3, 4{]}     \\
            Attention types         & {[}`F', `SF'{]} & {[}`F', `SF'{]} & {[}`F', `SF'{]}  & {[}`N', `F', `SF'{]} & {[}`F', `SF'{]} & {[}`N', `F', `SF'{]} & {[}`F', `SF'{]} & {[}`F', `SF'{]} \\
            Attention heads         & 4              & 6              & 4               & 4                   & 4              & 4                   & 5              & 4              \\
                                    &                &                &                 &                     &                &                     &                &                \\
            \textbf{Decoder}                 &                &                &                 &                     &                &                     &                &                \\\midrule
            Channels                & 1024           & 1024           & 1024            & 1024                & 1024           & 1024                & 1280           & 1536           \\
            Attention heads         & 8              & 8              & 8               & 8                   & 8              & 8                   & 10             & 12             \\
            Blocks (channels)  & 3              & 3              & 3               & 3                   & 3              & 3                   & 3              & 3              \\
            Blocks (positions) & 5              & 5              & 5               & 5                   & 5              & 5                   & 5              & 5              \\
            Blocks (values)    & 5              & 5              & 5               & 5                   & 5              & 5                   & 5              & 5              \\
                                    &                &                &                 &                     &                &                     &                &                \\
            \textbf{Training}                              &                &                &                 &                     &                &                     &                &                \\\midrule
            Dropout rate            & 0.05           & 0.05           & 0.05            & 0.05                & 0.05           & 0.25                & 0.05           & 0.             \\
            Training steps     & 2e5            & 3e5            & 2e5             & 3e5                 & 2e5            & 0.5e5               & 3e5            & 1.2e5          \\
            Parameters              & 470M           & 662M           & 470M            & 435M                & 470M           & 435M                & 711M           & 829M           \\ \bottomrule
        \end{tabular}
    }
    \label{tab:hyperparams}
\end{table*}

\section{Ablation studies}
\label{sec:ablations}
To investigate the impact of our model design choices we trained a selection of models on Kinetics600 video modelling at $64\times64$ resolution, while varying key hyperparameters. To reduce the computational cost of the ablations, we use a reduced size model, with 256 U-Net base channels, 256 decoder channels, a [2, 3] U-Net block structure, and 4 attention heads for both the U-Net and Transformer decoder. This smaller model has approximately 54M parameters, depending on the ablated configuration. We train the models for 100000 steps, which we found to be sufficient to characterize impactful hyperparameter choices. Otherwise we use the same settings as listed in the K600 column in Table \ref{tab:hyperparams}. We use FVD and negative log-likelihood - reported as bits-per-dimension (BPD) - as our evaluation metrics. The negative log-likelihood tracks the performance of our model when predicting each variable in the DCT sequence data. It is the objective we optimize, and can provide a clearer indication of the impact of design and hyperparameter choices than sample-based metrics. 

\paragraph{DCT quality settings.} First we investigate the impact of the DCT quality settings we use when converting images to DCT sequences. These settings affect the visual fidelity of the training data, with lower settings producing visible artifacts in image-reconstructions. Higher settings are essentially indistinguishable to the human eye from the original images. 

In particular we vary the DCT quantization quality, and the use of chroma downsampling. DCT quantization quality refers to the quantization level used in DCT block compression, following the JPEG standard paramterization of quantization matrices~\cite{DBLP:journals/cacm/Wallace91}. Chroma downsampling refers to the case where the chroma image channels are spatially downsampled 2x before applying block DCT compression. This is a widely used technique that enables higher levels of image compression while preserving perceptual quality to a large extent.
\begin{table}[t]
    \centering
    \caption{Model performance by DCT quality setting. FVD$^{\text{rec}}$ referers to FVD computed on video reconstructions from the validation set.}
    \begin{tabularx}{0.6\textwidth}{YYYY}\toprule
        DCT & Chroma & & \\
        Quality&Downsampling & FVD$\downarrow$ &FVD$^{\text{rec}}$$\downarrow$ \\\midrule
        50 & yes &437.4 &199.6 \\
        72 & yes &338.7 &75.8\\
        95 & yes &181.2 &5.5\\
        95 & no &184.2 &1.6\\
        99 & no &219.3 &0.1\\\bottomrule
    \end{tabularx}
    \label{tab:dct_setting_ablation}
\end{table}
In Table \ref{tab:dct_setting_ablation} we show the resulting performance metrics for models trained with the various quality settings. We don't report BPD, as it isn't comparable as a performance metric across different compression levels. We report FVD scores for both trained models and for reconstructions of ground truth videos to show how sensitive the metric is to the compression settings. 

For reconstructed  videos, we find that FVD is highly sensitive to DCT quality settings, which motivated our use of higher settings in most experiments. For trained models however, we see FVD scores initially decreasing with improved quality settings, but then increasing as we get to the highest settings. This may be because the modelling task is more challenging at higher quality settings, and model capacity is pushed to a greater extent. In general the FVD scores for trained models are generally higher than the reconstructions, as outputs are sampled and don't match the ground truth perfectly.

\paragraph{Number of context frames.} To test the extent to which our model benefits from conditioning on additional context frames, we train models with between 1-5 context video frames. Table \ref{tab:context_frames_ablation} shows that performance increases monotonically with the number of context frames, with a large drop in FVD going from 1-2 context frames. The number of context frames also impacts the training speed, with 5-frame models training 2-3$\times$ slower than a single context frame model. We anticipate that additional context frames will continue to yield performance improvements, at the cost of slower training, and increased memory consumption.  Whether this is the best use of a computational budget versus alternative model scaling approaches is an interesting question that we leave for future work.
\begin{table}[t]
    \centering
    \caption{Model performance by number of context frames.}
    \begin{tabularx}{0.55\textwidth}{YYYY}\toprule
        Context & & & Training \\
        frames &FVD$\downarrow$ & BPD$\downarrow$ & SPS$\uparrow$\\\midrule
        1 &258.51 &0.412 & 1.35\\
        2 &232.67 &0.393 & 1.16\\
        3 &218.14  &0.385 & 0.71 \\
        4 &211.20 &0.380 & 0.61\\
        5 &200.62 &0.377 & 0.54\\\bottomrule
    \end{tabularx}
    \label{tab:context_frames_ablation}
\end{table}

\paragraph{Multi-frame U-Net attention type.} We compare U-Net encoders trained with and without self-attention components. In our default configuration, we use attention layers to exchange information across input frames in the encoder, resulting in contextualized frame representations which are passed to the decoder (Section \ref{subsec:architecture}, Table \ref{tab:hyperparams}). We compare to a baseline model that omits self-attention layers, resulting in slightly smaller model with 46M parameters. We find that the baseline is about 19$\%$ faster, but the attention-enhanced model achieves substantially better overall performance.
\begin{table}[t]
    \centering
    \caption{Model performance with and without U-Net attention components.}
    \begin{tabularx}{0.55\textwidth}{YYYY}\toprule
        U-Net & & & Training \\
        Attention&FVD$\downarrow$ & BPD$\downarrow$ &  SPS$\uparrow$\\\midrule
        No &241.39 &0.389 & 0.64\\
        Yes &200.62 &0.377 &  0.54\\\bottomrule
    \end{tabularx}
    \label{tab:unet_attention_ablation}
\end{table}

\paragraph{Multi-query decoder attention.} We compare models trained with and without multi-query attention~\cite{shazeer2019} and compare performance and sampling speed for our models. Unlike standard multi-heads attention, multi-query attention shares key and value vectors across attention heads, while using distinct query heads. This substantially improves sampling speed for cache-based autoregressive sampling implementations, as it reduces the memory bandwidth required to access values from the cache. Consistent with Shazeer et al. \cite{shazeer2019}, we find that multi-query attention provides a 6.5$\times$ boost to sampling speeds, while achieving almost identical performance. Table \ref{tab:multi_query_ablation} shows these results.
\begin{table}[th]
    \centering
    \caption{Model performance with multi-query and multi-head attention.}
    \begin{tabularx}{0.6\textwidth}{YYYY}\toprule
        & & & Sample \\
        Attention &FVD$\downarrow$ & BPD$\downarrow$ & FPS$\uparrow$ \\\midrule
        Multi-head &200.98 &0.377 & 0.86\\
        Multi-query &200.62 &0.377 &  5.60\\\bottomrule
    \end{tabularx}
    \label{tab:multi_query_ablation}
\end{table}

\section{Objectron data pre-processing} 
\label{sec:objectron_data}
Objectron is a dataset of object-centric videos with camera and object pose annotations \cite{objectron2021}. For our view-synthesis model we take central crops from the portrait-orientation video frames, and resize to 192x192 resolution. We sub-sample the input videos with a stride of 10 to reduce temporal redundancy. We evaluate using the first 20 frames of each clip in the strided test set, using frame 4 for 1-view predictions, and frames 4 and 11 for 2-view predictions.

\section{Additional view synthesis results}
\label{sec:view_synth}
\begin{sidewaysfigure*}
    \centering
    \begin{tabular}{{>{\centering\arraybackslash} m{0.9\linewidth}m{0.1\linewidth}}}
         \includegraphics[width=\linewidth]{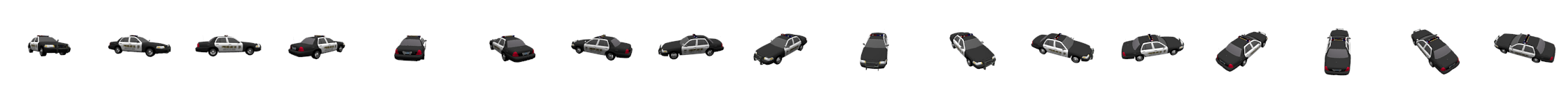} & Actual\\
         \includegraphics[width=\linewidth]{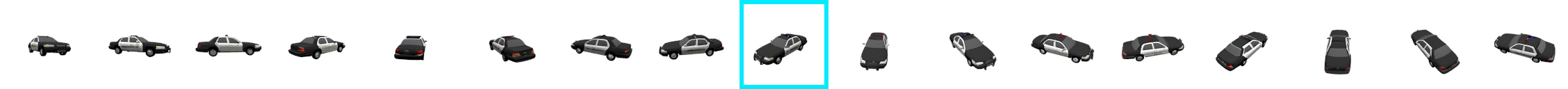} & Prediction\\
         \includegraphics[width=\linewidth]{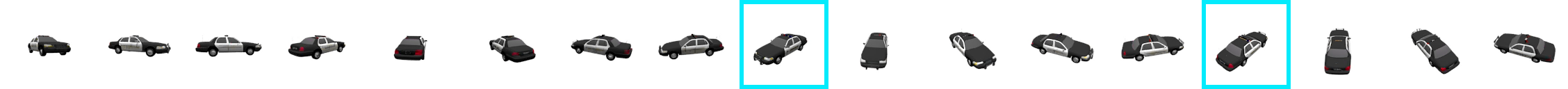} & Prediction\\
          \includegraphics[width=\linewidth]{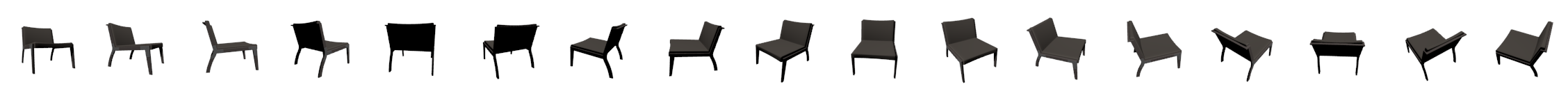} & Actual\\
         \includegraphics[width=\linewidth]{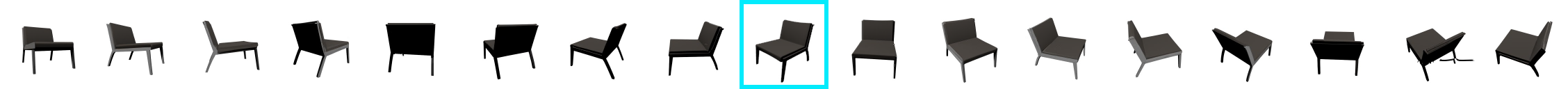} & Prediction\\
         \includegraphics[width=\linewidth]{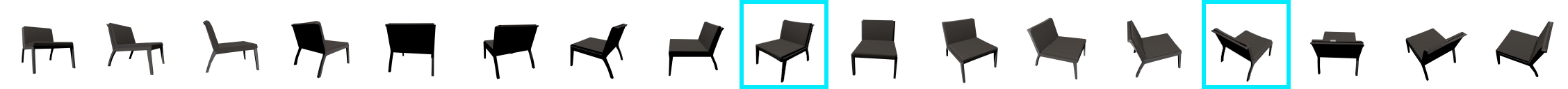} & Prediction\\
          \includegraphics[width=\linewidth]{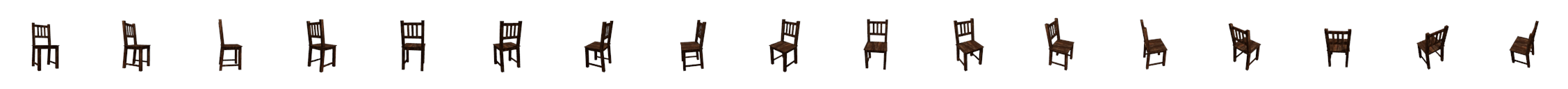} & Actual\\
         \includegraphics[width=\linewidth]{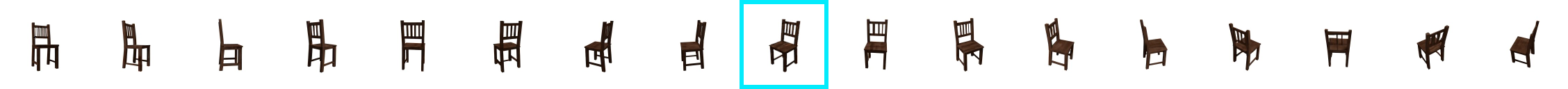} & Prediction\\
         \includegraphics[width=\linewidth]{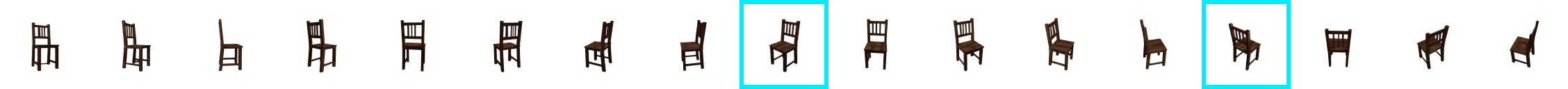} & Prediction\\
    \end{tabular}
    \caption{\textbf{Shapenet results.} For each video clip the top row shows ground truth, the middle row shows model outputs, and the bottom row shows prediction errors. For this dataset the model conditions on the actions of the robot arm and is therefore able to track the robot movement closely.}
    \label{fig:shapenet}
\end{sidewaysfigure*}

\begin{sidewaysfigure*}
    \centering
    \begin{tabular}{{>{\centering\arraybackslash} m{0.9\linewidth}m{0.1\linewidth}}}
         \includegraphics[width=\linewidth]{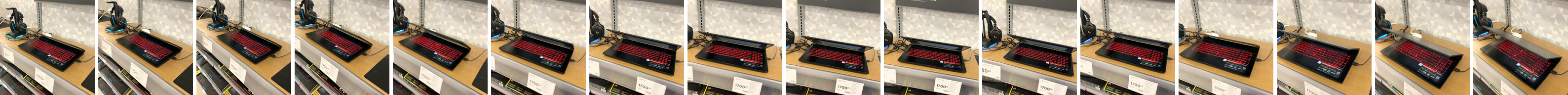} & Actual\\
         \includegraphics[width=\linewidth]{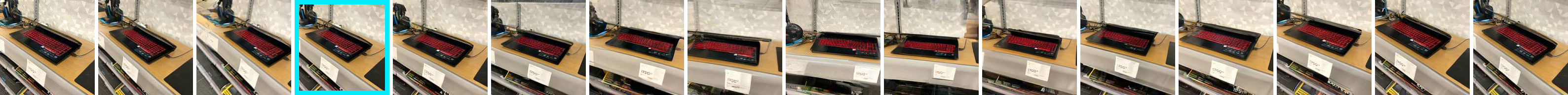} & Prediction\\
         \includegraphics[width=\linewidth]{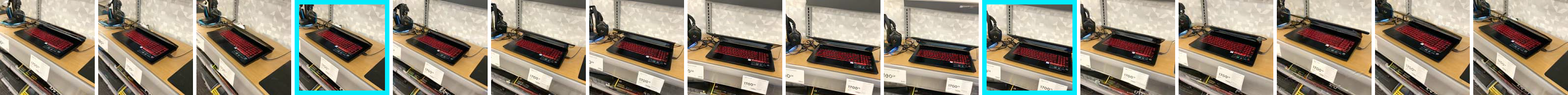} & Prediction\\
          \includegraphics[width=\linewidth]{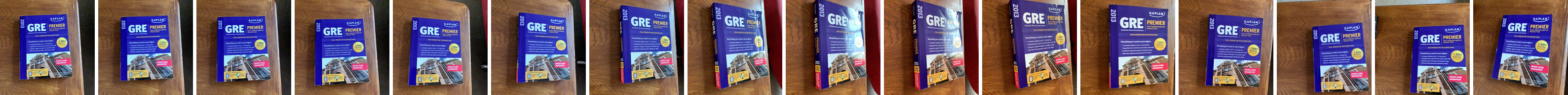} & Actual\\
         \includegraphics[width=\linewidth]{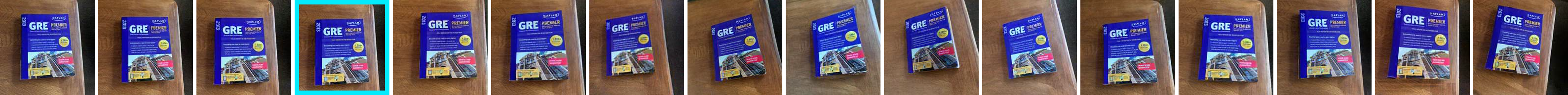} & Prediction\\
         \includegraphics[width=\linewidth]{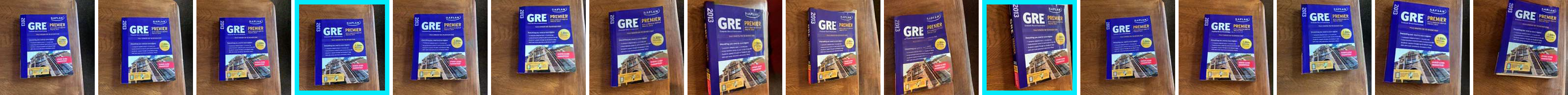} & Prediction\\
          \includegraphics[width=\linewidth]{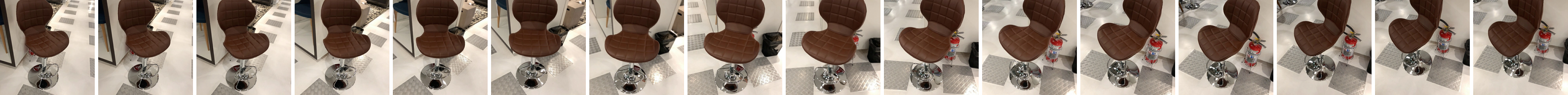} & Actual\\
         \includegraphics[width=\linewidth]{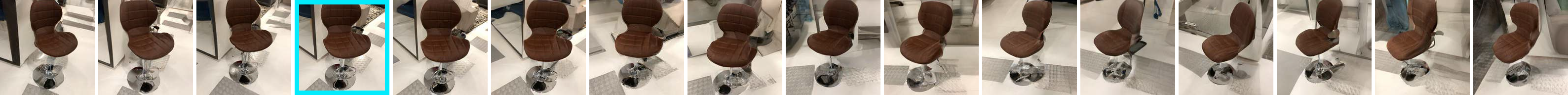} & Prediction\\
         \includegraphics[width=\linewidth]{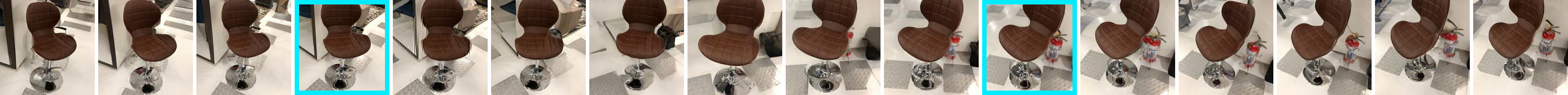} & Prediction\\
    \end{tabular}
    \caption{\textbf{Objectron results.} For each video clip the top row shows ground truth, the middle row shows model outputs, and the bottom row shows prediction errors. For this dataset the model conditions on the actions of the robot arm and is therefore able to track the robot movement closely.}
    \label{fig:objectron}
\end{sidewaysfigure*}

Videos ``ShapeNet (1 context view, 128x128)", ``ShapeNet (2 context views, 128x128)", ``Objectron (1 context view, 192x192)" and ``Objectron (2 context views, 192x192)" at \url{https://sites.google.com/view/transframer}\linebreak show test-set novel view synthesis outputs for ShapeNet and Objectron datasets, for 1 and 2 input views. Figures \ref{fig:shapenet} and \ref{fig:objectron} also show individual video frames for a reduced selection of scenes. For both videos and images the context frames are highlighted in Blue. 

\paragraph{Objectron results} Table \ref{tab:objectron} shows sample metric results for Objectron, for 1 and 2 input views. As there are no existing models that evaluate on this dataset, we cannot easily  compare to alternative methods. However our results may be of interest as a comparison for future work.

\begin{table}[th]
    \centering
    \caption{View synthesis results on Objectron for 1 and 2 input views. FID refers to the per-scene FID described in Section 4.2 in the main paper.}
    \begin{tabular}{lcccc}\toprule
        \textbf{Objectron} &FID$\downarrow$ &PSNR$\uparrow$ &SSIM$\uparrow$ &LPIPS$\downarrow$ \\\midrule
        1 view & 150.03 & 14.14 & 0.34 & 0.41 \\
        2 view &  109.29& 13.33 & 0.33 & 0.35\\
        \bottomrule
    \end{tabular}
    \label{tab:objectron}
\end{table}

\section{Additional video generation results}
\label{sec:extra_videos}
\begin{sidewaysfigure*}
    \centering
    \begin{tabular}{{>{\centering\arraybackslash} m{0.925\linewidth}m{0.075\linewidth}}}
         \includegraphics[width=\linewidth]{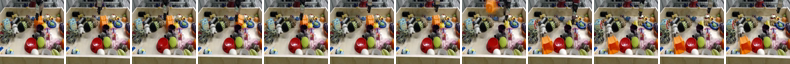} & Actual\\
         \includegraphics[width=\linewidth]{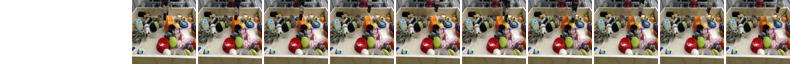} & Prediction\\
         \includegraphics[width=\linewidth]{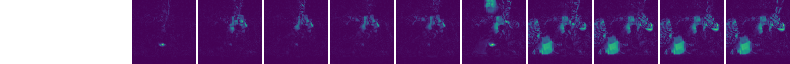} & Errors\\
         \includegraphics[width=\linewidth]{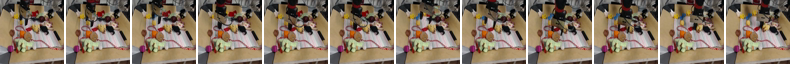} & Actual\\
         \includegraphics[width=\linewidth]{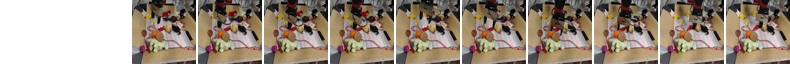} & Prediction \\
         \includegraphics[width=\linewidth]{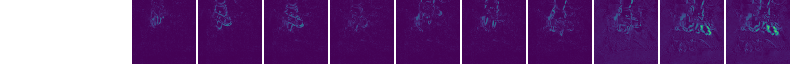} & Errors\\
    \end{tabular}
    \caption{\textbf{Robonet (64x64) results.} For each video clip the top row shows ground truth, the middle row shows model outputs, and the bottom row shows prediction errors. For this dataset the model conditions on the actions of the robot arm and is therefore able to track the robot movement closely.}
    \label{fig:robonet_64}
\end{sidewaysfigure*}

\begin{sidewaysfigure*}
    \centering
    \begin{tabular}{{>{\centering\arraybackslash} m{0.925\linewidth}m{0.075\linewidth}}}
         \includegraphics[width=\linewidth]{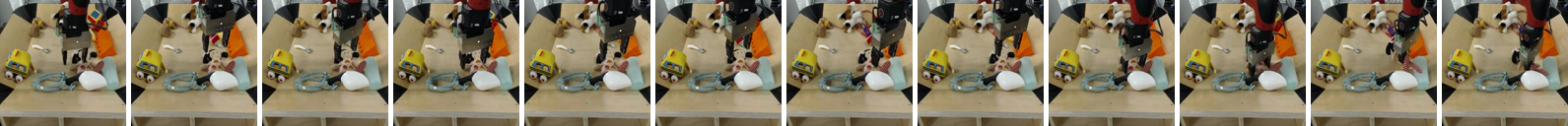} & Actual\\
         \includegraphics[width=\linewidth]{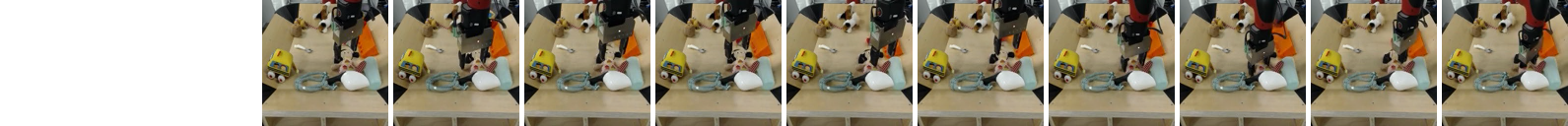} & Prediction\\
         \includegraphics[width=\linewidth]{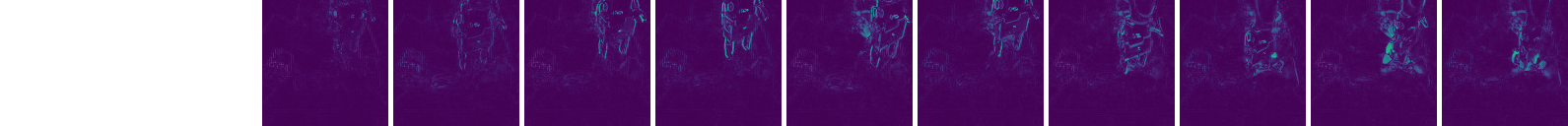} & Errors\\
         \includegraphics[width=\linewidth]{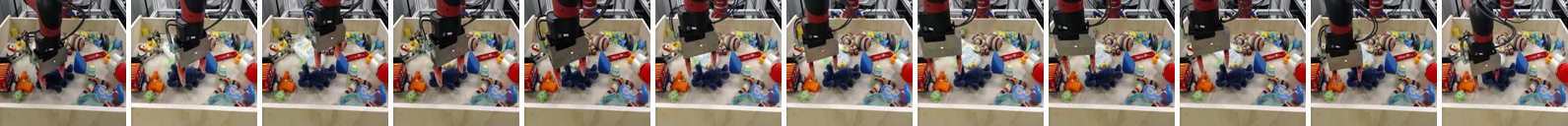} & Actual\\
         \includegraphics[width=\linewidth]{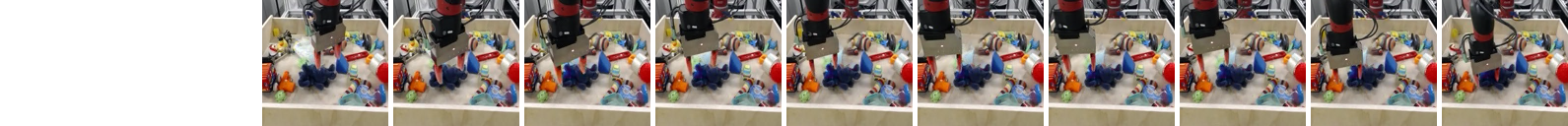} & Prediction \\
         \includegraphics[width=\linewidth]{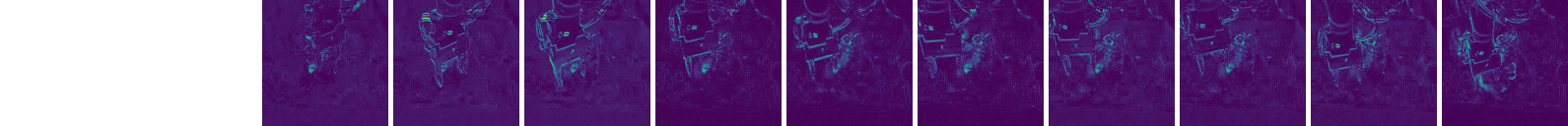} & Errors\\
    \end{tabular}
    \caption{\textbf{Robonet (128x128) results.} For each video clip the top row shows ground truth, the middle row shows model outputs, and the bottom row shows prediction errors. For this dataset the model conditions on the actions of the robot arm and is therefore able to track the robot movement closely.}
    \label{fig:robonet_128}
\end{sidewaysfigure*}

\begin{sidewaysfigure*}
    \centering
    \begin{tabular}{{>{\centering\arraybackslash} m{0.9\linewidth}m{0.1\linewidth}}}
         \includegraphics[width=\linewidth]{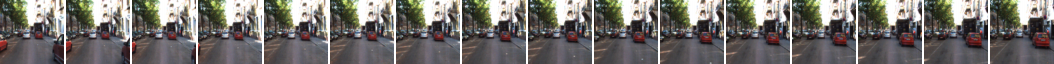} & Actual\\
         \includegraphics[width=\linewidth]{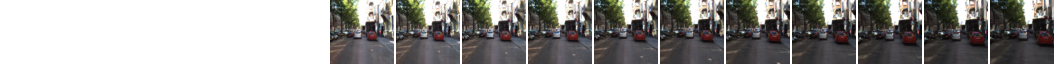} & Prediction\\
         \includegraphics[width=\linewidth]{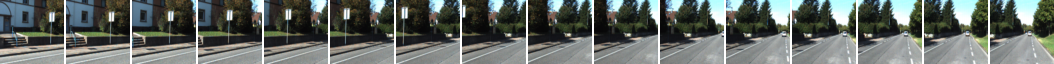} & Actual\\
         \includegraphics[width=\linewidth]{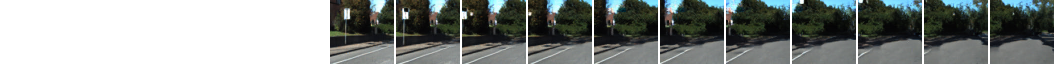} & Prediction\\
         \includegraphics[width=\linewidth]{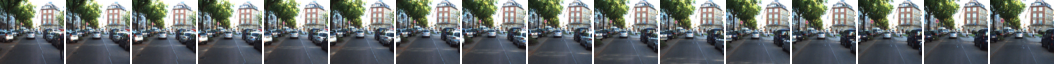} & Actual\\
         \includegraphics[width=\linewidth]{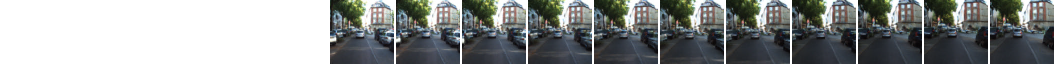} & Prediction\\
          \includegraphics[width=\linewidth]{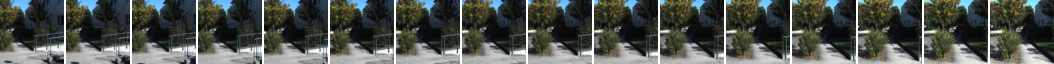} & Actual\\
         \includegraphics[width=\linewidth]{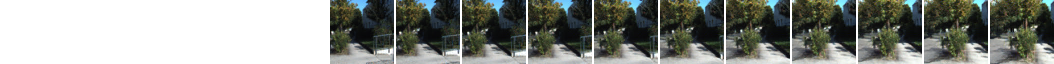} & Prediction\\
    \end{tabular}
    \caption{\textbf{Kitti results.} For each video clip the top row shows ground truth, the middle row shows model outputs, and the bottom row shows prediction errors. For this dataset the model conditions on the actions of the robot arm and is therefore able to track the robot movement closely. The KITTI figure (Figure \ref{fig:kitti} shows a truncated sample due to space constraints.}
    \label{fig:kitti}
\end{sidewaysfigure*}

Videos ``RoboNet (64x64)", ``RoboNet (128x128)" and ``KITTI (64x64)" at \url{https://sites.google.com/view/transframer} show test-set video generation outputs for RoboNet (64x64), RoboNet (128x128), and KITTI datasets respectively. Orange and green frame outlines indicate context and predicted frames respectively. Figures \ref{fig:robonet_64}, \ref{fig:robonet_128} and \ref{fig:kitti} also show individual video frames for a reduced selection of scenes, but we recommend the videos for viewing clarity. For the RoboNet examples we visualize errors between the predicted and true target frames, as the action-conditioning results in a close predictive match. Video \texttt{\small long\_video.webm} six videos conditioned on a single image with an unroll of about 30 seconds. Two of the videos are conditioned on the same image, showing diversity in generations. Most of the conditioning images were personal photos taken by the authors, but the lower middle is under the \href{https://creativecommons.org/licenses/by/2.5/deed.en}{CC-BY 2.5} license and created by \texttt{\small Onofre\_Bouvila} on Wikipedia. It is available at this \href{https://en.wikipedia.org/wiki/Street#/media/File:Typical_Street_In_The_Royal_Borough_Of_Kensington_And_Chelsea_In_London.jpg}{link}

\paragraph{Robonet (128x128) results} Table \ref{tab:metrics_robonet_128} shows sample metrics for Transframer on RoboNet (128x128). We use the same evaluation protocol as the 64x64 resolution data: 2 context frames and 10 rollout frames on the 256 clips in the test set specified by Babaeizadeh et al.\ \cite{DBLP:journals/corr/abs-2106-13195}. For each clip we generate 100 predictions. For the PSNR, SSIM, and LPIPS metrics we choose the best prediction in each case by PSNR score, and for FVD we average over the FVD scores for each prediction.

\begin{table}[t]
    \centering
    \caption{Sample metrics for generated videos on action-conditional RoboNet at 128x128 resolution. We use 2 context frames and generate 10 frames.}
    \begin{tabular}{lcccc}\toprule
        \textbf{RoboNet (128x128)}&FVD$\downarrow$ &PSNR$\uparrow$ &SSIM$\uparrow$ &LPIPS$\downarrow$ \\\midrule
        \modelname &63.1 &28.5 &90.8 &0.030 \\
        \bottomrule
    \end{tabular}
    \label{tab:metrics_robonet_128}
\end{table}

\section{Additional multi-task results and details}
\label{sec:multi}

We trained a single multi-task model jointly on the 8 tasks/datasets listed below. In all cases we used 256x256 resolution by random cropping, resizing and padding the images in training and resizing in validation. 

We show 5 samples obtained using high temperature for 2 examples from each task/dataset in figs.~\ref{fig:multitask1} and \ref{fig:multitask2}. 

\paragraph{ImageNet classification~\cite{deng2009imagenet}.} We recast ImageNet classification as frame prediction by using one-hot images as targets. One-hot images are 2D analogues of the one-hot vectors traditionally used for classification: each class is encoded as a white 8x8 patch on a black background. The patches for different classes do not overlap. This configuration allows us to fit the targets for all 1000 classes on a 256x256 image. On ImageNet, the multi-task model gets 48\% top-1 accuracy, while the same model trained on single-task classification gets 69\%. The results already suggest that the model can perform object recognition to reasonable levels but also that the multi-task case is more challenging as the model needs to accommodate a wide range of predictions. We also trained single-task models to predict images showing the target class name rendered in plain text. These models trained much  slower, achieving 7\% top-1 accuracy after the same number of epochs as the equivalent model trained using one-hot images (69\%). 

\paragraph{Cityscapes semantic segmentation~\cite{cordts2016cityscapes}.} Cityscapes is a popular dataset for segmentation of street scenes. We render the ground truth segmentation maps using the Cityscapes colormap. The multi-task model gets around 28\% mean IoU when trained on the fine-grained training set and evaluating on the validation set (256x256 resolution). The results look quite decent visually though.

\paragraph{PASCAL VOC semantic segmentation~\cite{everingham2010pascal}.} This is a classic semantic segmentation task with 20 object categories. We render the ground truth outputs using the iconic PASCAL VOC colormap.

\paragraph{COCO object detection~\cite{lin2014microsoft}.} We render all ground truth boxes for an image on a white background. The color of the box is determined by the object category. We also used the PASCAL VOC colormap here. Results seems to reflect the overall complexity of the original sets of boxes. It would be interesting to compare numerically the difference in mean squared error between our rendered box images and those produced using state-of-the-art object detectors, we leave this to future work.

\paragraph{COCO scene parsing.} This is a non-standard task where we use the COCO panoptic annotations (object and background regions) and render each region using the average of RGB values inside it -- no class labels are used here. The model seems quite good at this task, suggesting segmentation skills are superior to its object recognition skills. 

\paragraph{Nyu-Depth~\cite{Silberman:ECCV12}.} Nyu Depth is a single-image depth estimation benchmark. We render depth using the Viridis matplotlib colormap. Results look visually coherent, and multiple samples reflect to some extent the uncertainty in the task.

\paragraph{Kinetics600~\cite{carreira2018short} optical flow} This is another non-standard task where the goal is to predict optical flow from a single image. As "ground truth" optical flow we adopt the output of the TV-L1 algorithm~\cite{zach2007duality}. The model seems to implicitly capture the notion of humans as flow tends to align with human boundaries.

\paragraph{YouTube8M~\cite{abu2016youtube} future frame prediction.} We sample two frames 1 second apart from videos in the YouTube8M dataset. The model is trained to predict the future frame. This is an extremely challenging task as this dataset has in the wild videos from youtube, with fast motion and huge variety.

When training we sample dataset examples with the following probabilities:
\begin{table}[h]
    \centering
    \resizebox{0.95\linewidth}{!}{
        \begin{tabularx}{\linewidth}{YYYYYYYY}
            \toprule
            Pascal & 
            Cityscapes & 
            NYU depth & 
            COCO scene parsing & 
            COCO detection & 
            Imagenet & 
            Kinetics flow pred & 
            YT8M \\\midrule
            0.05 & 0.05 & 0.05 & 0.15 & 0.20 & 0.30 & 0.10 & 0.10 \\\bottomrule
        \end{tabularx}
    }
\end{table}

\renewcommand{\imwidthplot}{2.0cm}
\renewcommand{\viewtabletitlesize}{\small}
\begin{figure*}[t]
    \centering
        \setlength\tabcolsep{0.075cm}
        \resizebox{\linewidth}{!}{
        \begin{tabular}{cc|cccc|c}
        & \multicolumn{1}{c}{\viewtabletitlesize\underline{Context}} &  \multicolumn{4}{c}{\viewtabletitlesize\underline{Predictions}} & {\viewtabletitlesize\underline{Actual}} \\
          \vertlabel{Semantic Segm. \\ (Cityscapes)} & \includegraphics[width=\imwidthplot]{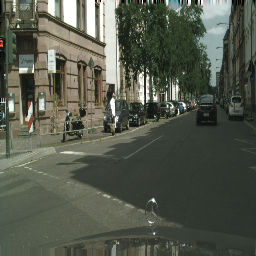} & \includegraphics[width=\imwidthplot]{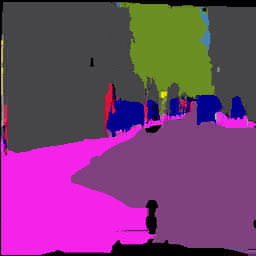} &  \includegraphics[width=\imwidthplot]{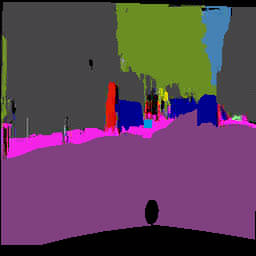} &
          \includegraphics[width=\imwidthplot]{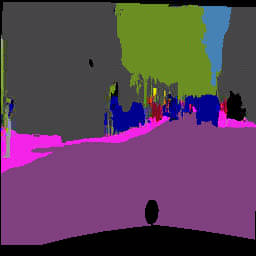} &
          \includegraphics[width=\imwidthplot]{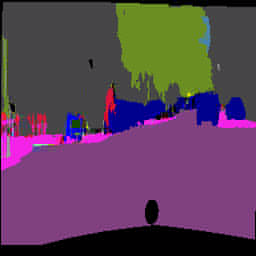} &
          \includegraphics[width=\imwidthplot]{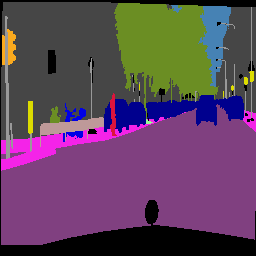} \\
          \vertlabel{Semantic Segm. \\ (PASCAL)} & \includegraphics[width=\imwidthplot]{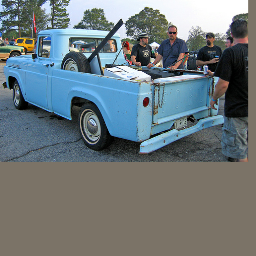} 
          & \includegraphics[width=\imwidthplot]{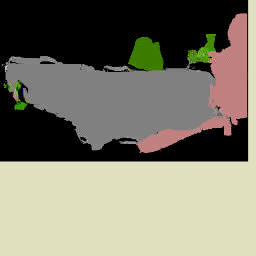} & 
          \includegraphics[width=\imwidthplot]{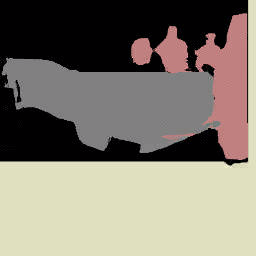} & \includegraphics[width=\imwidthplot]{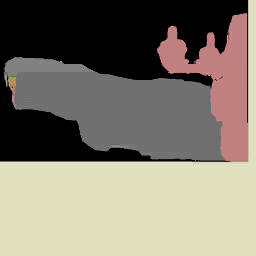} &
          \includegraphics[width=\imwidthplot]{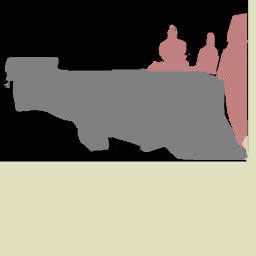} &
          \includegraphics[width=\imwidthplot]{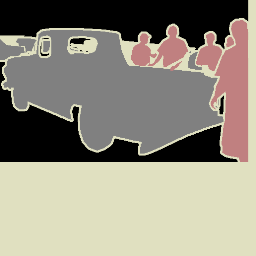} \\ \vertlabel{Detection \\ (COCO)} &  \includegraphics[width=\imwidthplot]{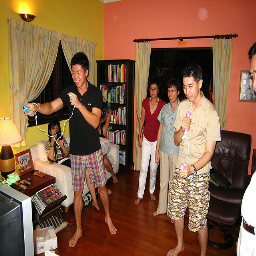}  &
          \includegraphics[width=\imwidthplot]{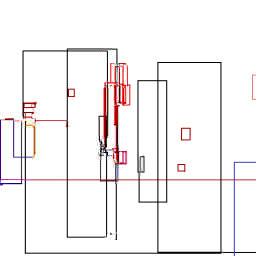} & \includegraphics[width=\imwidthplot]{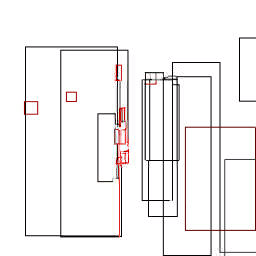} & 
          \includegraphics[width=\imwidthplot]{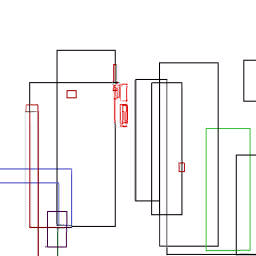} 
          & \includegraphics[width=\imwidthplot]{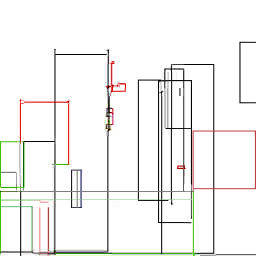} &
          \includegraphics[width=\imwidthplot]{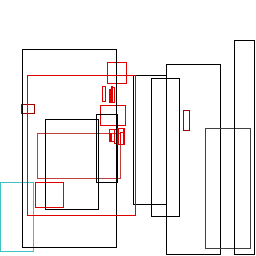} \\ \vertlabel{Scene parsing \\ (COCO)} & \includegraphics[width=\imwidthplot]{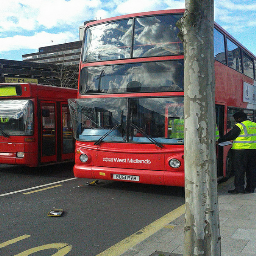} 
          & \includegraphics[width=\imwidthplot]{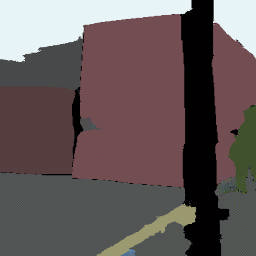} & 
          \includegraphics[width=\imwidthplot]{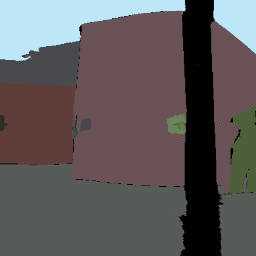} & \includegraphics[width=\imwidthplot]{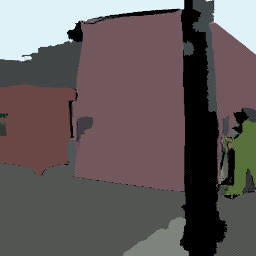} &
          \includegraphics[width=\imwidthplot]{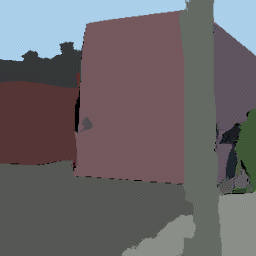} &
          \includegraphics[width=\imwidthplot]{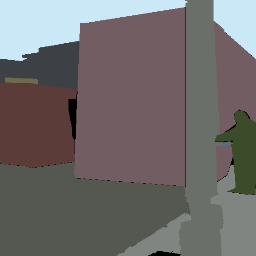} \\
        \vertlabel{Classification \\ (ImageNet)} & \includegraphics[width=\imwidthplot]{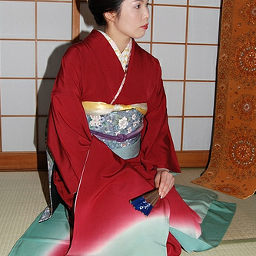}  &
          \includegraphics[width=\imwidthplot]{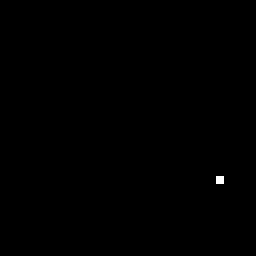} & \includegraphics[width=\imwidthplot]{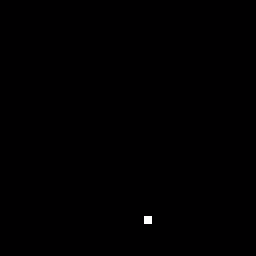} & 
          \includegraphics[width=\imwidthplot]{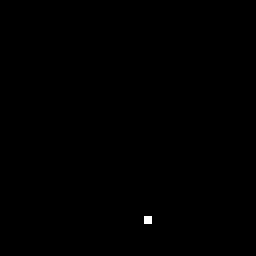} & \includegraphics[width=\imwidthplot]{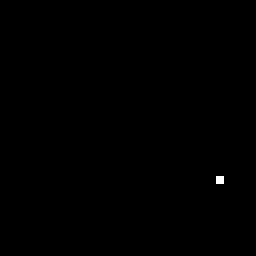} &
          \includegraphics[width=\imwidthplot]{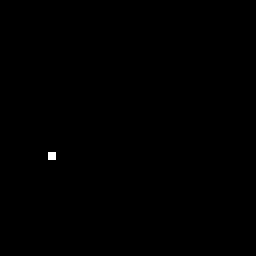} \\
        \vertlabel{Flow prediction \\ (Kinetics600)} & \includegraphics[width=\imwidthplot]{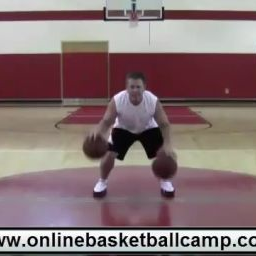}  &
          \includegraphics[width=\imwidthplot]{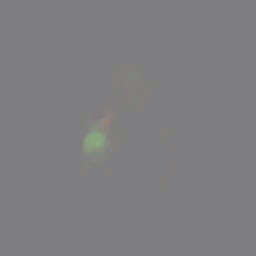} & \includegraphics[width=\imwidthplot]{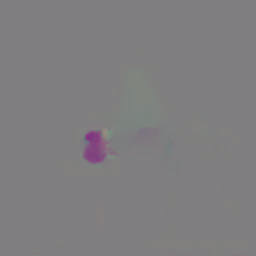} & 
          \includegraphics[width=\imwidthplot]{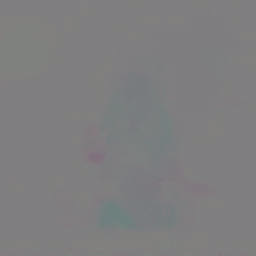} & \includegraphics[width=\imwidthplot]{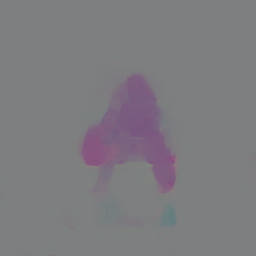} &
          \includegraphics[width=\imwidthplot]{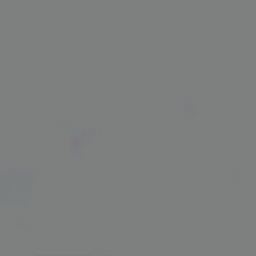} \\
        \vertlabel{Video modelling \\ (Youtube8M)} & \includegraphics[width=\imwidthplot]{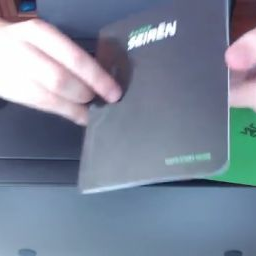}
          & \includegraphics[width=\imwidthplot]{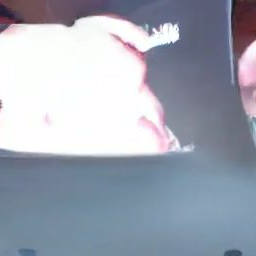} & 
          \includegraphics[width=\imwidthplot]{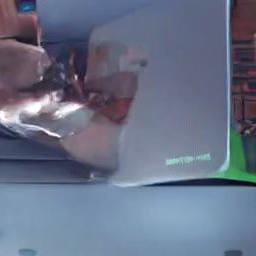} & \includegraphics[width=\imwidthplot]{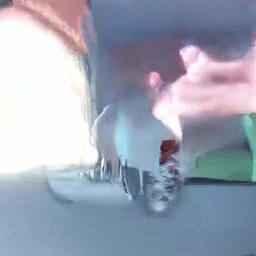} &
          \includegraphics[width=\imwidthplot]{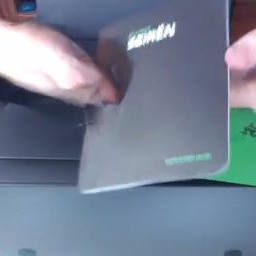} &
          \includegraphics[width=\imwidthplot]{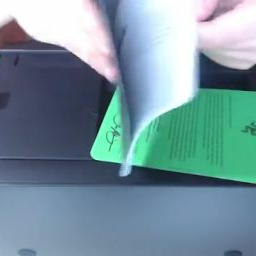} \\
        \vertlabel{Depth \\ estimation \\ (NYUDepth)} & \includegraphics[width=\imwidthplot]{graphics/multitask/supp/nyudepth_0_img.png} 
          & \includegraphics[width=\imwidthplot]{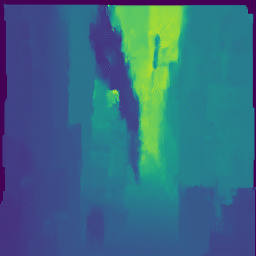} & 
          \includegraphics[width=\imwidthplot]{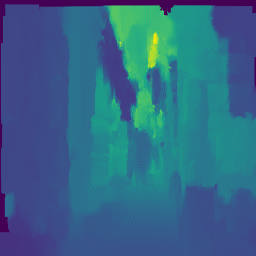} & \includegraphics[width=\imwidthplot]{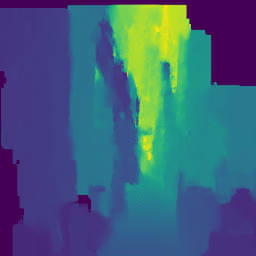} &
          \includegraphics[width=\imwidthplot]{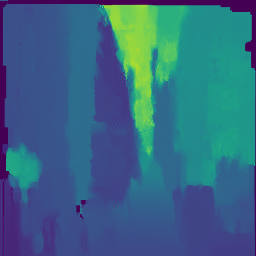} &
          \includegraphics[width=\imwidthplot]{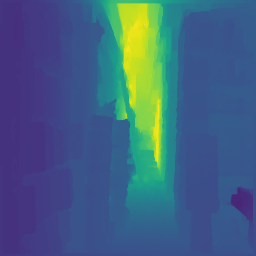} \\
        \end{tabular}
    }
    \caption{Example samples on the validation set of our multi-task setup, obtained using a single model. The model has no task specific parameters; in all cases making frame predictions conditioned on a context frame and a one-hot task embedding vector.}
    \label{fig:multitask1}
\end{figure*}

\renewcommand{\imwidthplot}{2.0cm}
\renewcommand{\viewtabletitlesize}{\small}
\begin{figure*}[t]
    \centering
        \setlength\tabcolsep{0.075cm}
        \resizebox{\linewidth}{!}{
        \begin{tabular}{cc|cccc|c}
        & \multicolumn{1}{c}{\viewtabletitlesize\underline{Context}} &  \multicolumn{4}{c}{\viewtabletitlesize\underline{Predictions}} & {\viewtabletitlesize\underline{Actual}} \\
          \vertlabel{Semantic Segm. \\ (Cityscapes)} & \includegraphics[width=\imwidthplot]{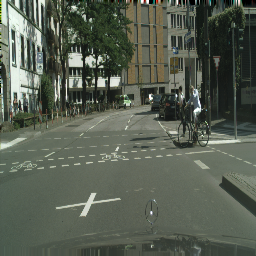} & \includegraphics[width=\imwidthplot]{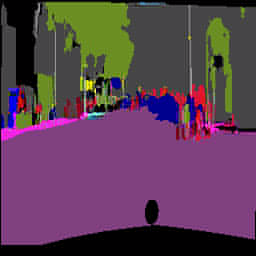} &  \includegraphics[width=\imwidthplot]{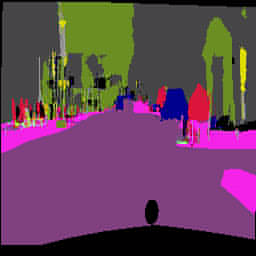} &
          \includegraphics[width=\imwidthplot]{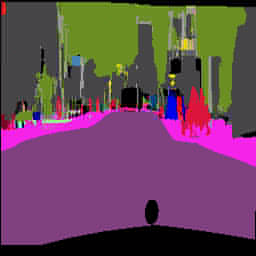} &
          \includegraphics[width=\imwidthplot]{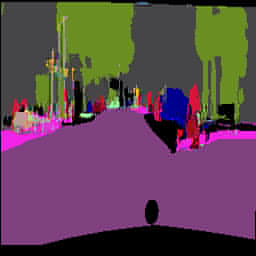} &
          \includegraphics[width=\imwidthplot]{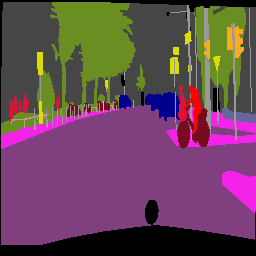} \\
          \vertlabel{Semantic Segm. \\ (PASCAL)} & \includegraphics[width=\imwidthplot]{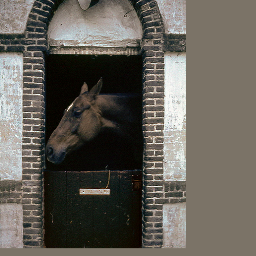} & \includegraphics[width=\imwidthplot]{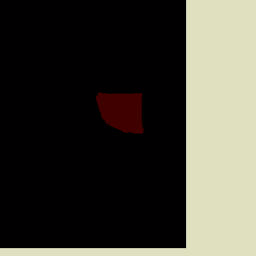} & \includegraphics[width=\imwidthplot]{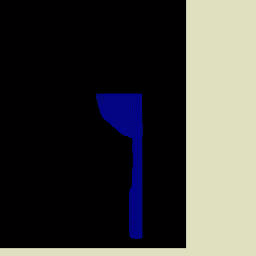} & 
          \includegraphics[width=\imwidthplot]{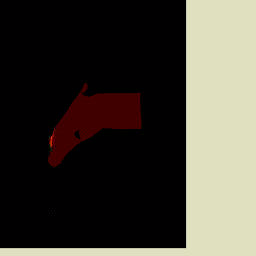} & \includegraphics[width=\imwidthplot]{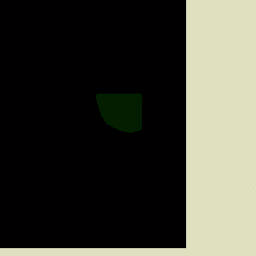} &
          \includegraphics[width=\imwidthplot]{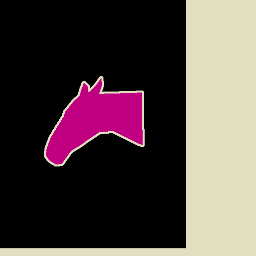} \\ \vertlabel{Detection \\ (COCO)} &  \includegraphics[width=\imwidthplot]{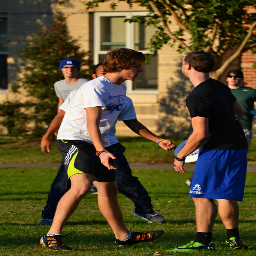}  &
          \includegraphics[width=\imwidthplot]{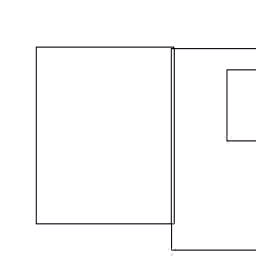} & 
          \includegraphics[width=\imwidthplot]{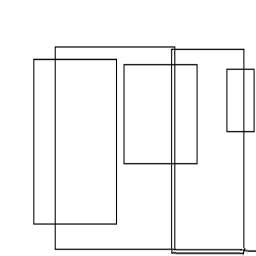} & \includegraphics[width=\imwidthplot]{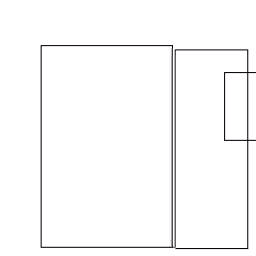} &
          \includegraphics[width=\imwidthplot]{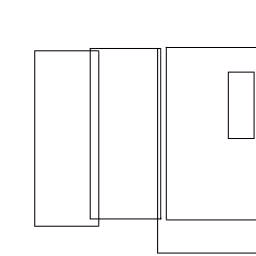} &
          \includegraphics[width=\imwidthplot]{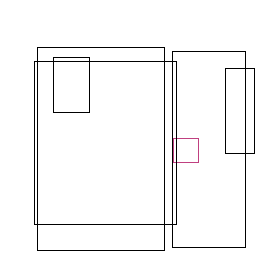} \\ \vertlabel{Scene parsing \\ (COCO)} & \includegraphics[width=\imwidthplot]{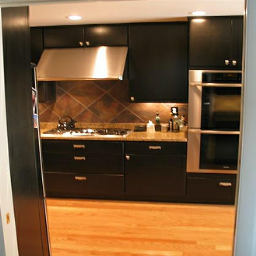}  &
          \includegraphics[width=\imwidthplot]{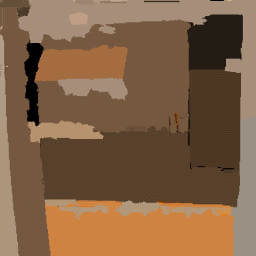} & \includegraphics[width=\imwidthplot]{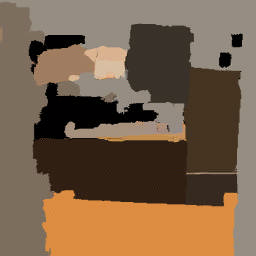} & 
          \includegraphics[width=\imwidthplot]{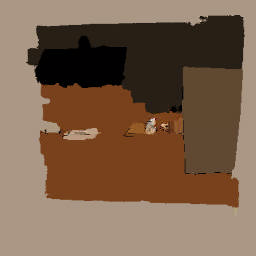} & \includegraphics[width=\imwidthplot]{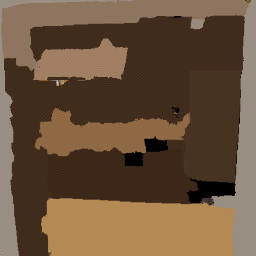} &
          \includegraphics[width=\imwidthplot]{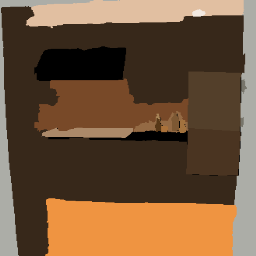} \\
        \vertlabel{Classification \\ (ImageNet)} & \includegraphics[width=\imwidthplot]{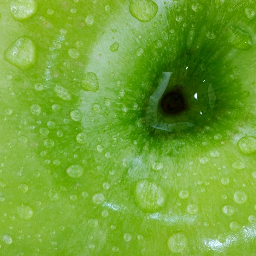}  &
          \includegraphics[width=\imwidthplot]{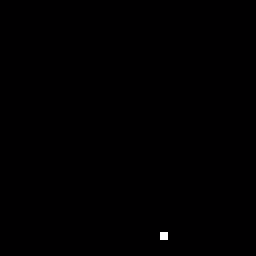} & \includegraphics[width=\imwidthplot]{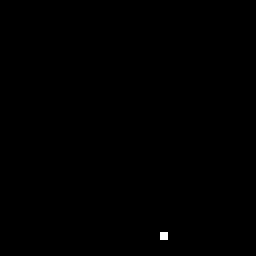} & 
          \includegraphics[width=\imwidthplot]{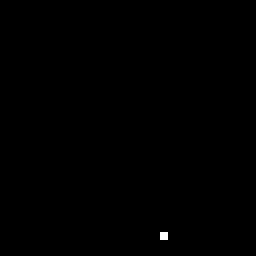} & \includegraphics[width=\imwidthplot]{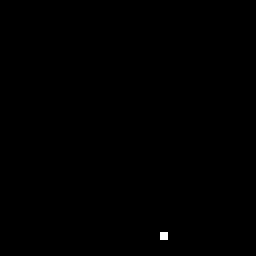} &
          \includegraphics[width=\imwidthplot]{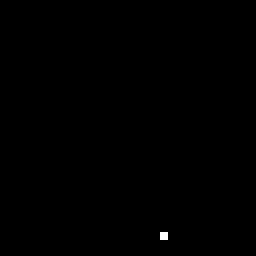} \\
        \vertlabel{Flow prediction \\ (Kinetics600)} & \includegraphics[width=\imwidthplot]{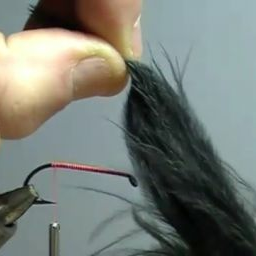}  &
          \includegraphics[width=\imwidthplot]{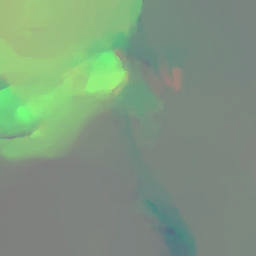} & \includegraphics[width=\imwidthplot]{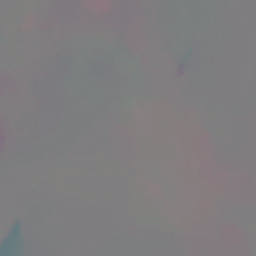} & 
          \includegraphics[width=\imwidthplot]{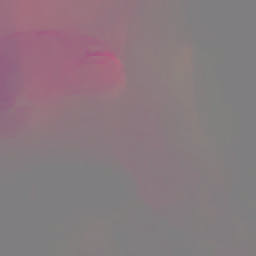} & \includegraphics[width=\imwidthplot]{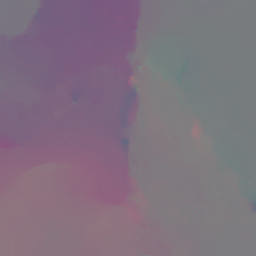} &
          \includegraphics[width=\imwidthplot]{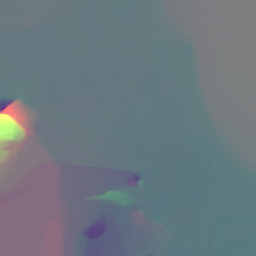} \\
        \vertlabel{Video modelling \\ (Youtube8M)} & \includegraphics[width=\imwidthplot]{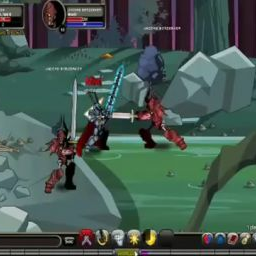}  &
          \includegraphics[width=\imwidthplot]{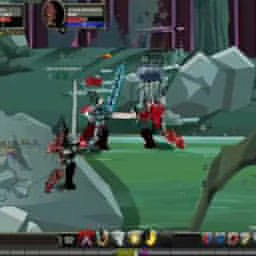} & \includegraphics[width=\imwidthplot]{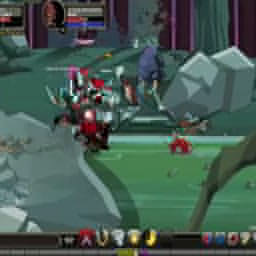} & 
          \includegraphics[width=\imwidthplot]{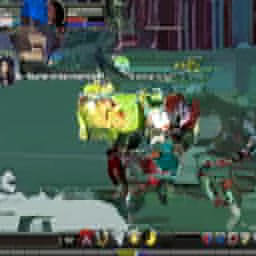} & \includegraphics[width=\imwidthplot]{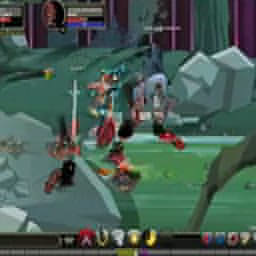} &
          \includegraphics[width=\imwidthplot]{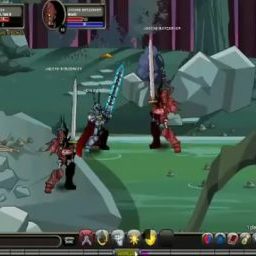} \\
        \vertlabel{Depth \\ estimation \\ (NYUDepth)} & \includegraphics[width=\imwidthplot]{graphics/multitask/supp/nyudepth_1_img.png}  &
          \includegraphics[width=\imwidthplot]{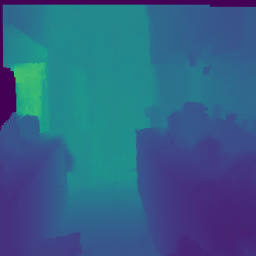} & \includegraphics[width=\imwidthplot]{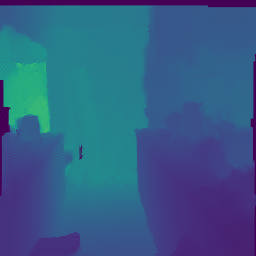} & 
          \includegraphics[width=\imwidthplot]{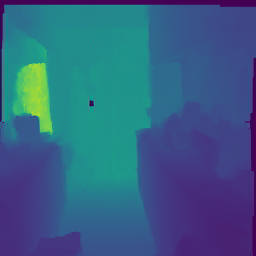} & \includegraphics[width=\imwidthplot]{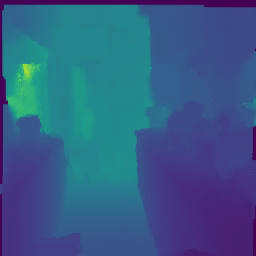} &
          \includegraphics[width=\imwidthplot]{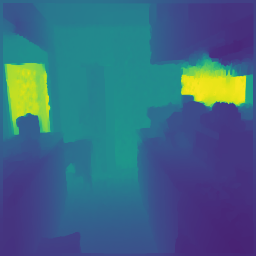} \\
        \end{tabular}
    }
    \caption{More example samples on the validation set of our multi-task setup, obtained using a single model.}
    \label{fig:multitask2}
\end{figure*}

\end{document}